\documentclass[font=12]{article}

\usepackage{arxiv}

\usepackage[utf8]{inputenc} 
\usepackage[T1]{fontenc}    
\usepackage[hidelinks]{hyperref}       
\usepackage{url}            
\usepackage{booktabs}       
\usepackage{algorithm}
\usepackage{algpseudocode}
\usepackage{threeparttable}

\usepackage{xcolor}

\usepackage{amsfonts}       
\usepackage{amsmath} 
\usepackage{amssymb}
\usepackage{nicefrac}       
\usepackage{microtype}      
\usepackage{bm}

\usepackage{lipsum}		
\usepackage{graphicx}
\usepackage{doi}
\usepackage{comment}
\usepackage{subcaption}

\usepackage[acronym]{glossaries}
\newacronym{cp}{CP}{conformal prediction}
\newacronym{cv}{CV}{cross-validation}
\newacronym{gp}{GP}{Gaussian process}
\newacronym{iid}{iid}{independent and identically distributed}
\newacronym{ml}{ML}{machine learning}
\newacronym{pca}{PCA}{principal component analysis}
\newacronym{pod}{POD}{proper orthogonal decomposition}
\newacronym{svd}{SVD}{singular value decomposition}
\newacronym{uq}{UQ}{uncertainty quantification}

\DeclareMathOperator*{\argmin}{arg\,min}

\title{Adaptive conformal prediction on the Grassmann manifold}
\title{Adaptive conformal prediction on the Grassmannian}
\date{} 					

\author{ \href{https://orcid.org/0000-0003-1264-1182}{\includegraphics[scale=0.06]{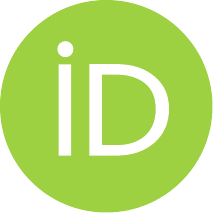}\hspace{1mm}Dimitrios Loukrezis}\\
	Scientific Computing Group\\
	Centrum Wiskunde \& Informatica\\
	Amsterdam, The Netherlands \\
	\texttt{d.loukrezis@cwi.nl} \\
    \And
    \href{XXX}{\includegraphics[scale=0.06]{orcid.pdf}\hspace{1mm}Zachary J. Grey} \\
Applied \& Computational Mathematics Division \\
 National Institute of Standards \& Technology \\
 Boulder, CO, USA \\
	\texttt{zachary.grey@nist.gov} \\
	\And
	\href{https://orcid.org/0000-0003-2272-2584}{\includegraphics[scale=0.06]{orcid.pdf}\hspace{1mm}Dimitris G.~Giovanis} \\
	Dpt. of Civil \& Systems Engineering\\
    Dpt. of Applied Mathematics and Statistics \\
	Johns Hopkins University\\
	Baltimore, MD, USA \\
	\texttt{dgiovan1@jhu.edu} \\
}

\renewcommand{\headeright}{}
\renewcommand{\undertitle}{}

\begin{document}
\maketitle

\begin{abstract}
We introduce a conformal prediction framework for regression with Grassmann-valued responses. 
The proposed method constructs prediction regions in the form of metric balls directly on the Grassmann manifold, using basis-invariant geometric distances as non-conformity measures and scaling the corresponding non-conformity scores through a learned prediction difficulty estimator. 
This yields locally adaptive, input-dependent prediction regions that respect the underlying manifold geometry, while retaining the finite-sample marginal coverage guarantee of split conformal prediction under exchangeability. 
We evaluate several Grassmannian distance metrics and compare their prediction-region efficiency on synthetic benchmarks and an airfoil inverse shape-design problem. 
The results highlight the advantages of geometry-aware non-conformity scores, while local adaptation improves empirical coverage uniformity across prediction difficulty levels and, in strongly heteroscedastic settings, can substantially reduce prediction-region size while maintaining the target coverage.
\end{abstract}


\section{Introduction}
\label{sec:intro}

\subsection{Motivation}
\label{sec:intro-motivation}
\Gls{cp} provides a distribution-free framework for \gls{uq} by complementing the point estimates of data-driven models with statistically calibrated prediction sets that quantify predictive uncertainty~\cite{vovk2005algorithmic,fontana2023conformal}. 
Notably, certain \gls{cp} variants offer marginal coverage guarantees, requiring the prediction set to contain a future response with a prescribed probability, often referred to as the target coverage. Recently, \gls{cp} has received much attention in scientific and engineering applications where \gls{ml} regression models are deployed as computationally efficient surrogates, yet reliable predictive \gls{uq} is required~\cite{wu2025sequential, jaber2025conformal, loukrezis2025conformalized, gopakumar2026uncertainty, hatstatt2026conformal, mabtoul2026uncertainty, stent2026conformal}. 

Conformal regression methods are generally formulated for scalar or Euclidean-valued responses, as these are also the most common observable outputs.
Discrepancies between predictions and observations are then measured using standard norms, and prediction sets take the form of intervals or Euclidean regions. 
However, if the response is instead a subspace or other geometric object, Euclidean prediction regions constructed from arbitrary coordinate or basis representations need not respect the geometry nor the invariances of the response space. 
This distinction is particularly important for models whose outputs admit low-dimensional geometric representations. 
For example, high-dimensional simulation data are frequently compressed through nonlinear dimension reduction or manifold learning before surrogate models are constructed~\cite{kalogeris2020diffusion, lataniotis2020extending, kontolati2022manifold, kontolati2022survey, soize2022probabilistic, giovanis2024polynomial, giovanis2026enabling}.
In these settings, physically equivalent representations may differ only through a change of basis (e.g., a rotational alignment) and must therefore be treated as the same geometric object.
Euclidean prediction regions that depend on a selected basis representation may fail to respect the invariances that define the underlying quantity of interest.
Therefore, uncertainty should be quantified in terms of variations between subspaces. 

In this work, we are interested in geometric responses (subspaces) residing on the Grassmann manifold (also referred to as the Grassmannian)~\cite{absil2004riemannian, bendokat2024grassmann}, denoted $\mathrm{Gr}(p, n)$ and formally defined as the set of all $p$-dimensional linear subspaces in $\mathbb{R}^n$. 
A subspace on the Grassmann manifold is commonly represented by an orthonormal basis matrix. 
However, this representation is not unique, as any rotation or reflection of the basis vectors spans the exact same subspace.
Thus, a single geometric point on the Grassmannian is formally understood as the equivalence class of all such orthonormal matrices.
This invariance arises naturally in \gls{pca} and \gls{pod}, where reduced-order bases are defined only up to transformations that preserve their span. 
It also extends to geometric representations in which shapes are encoded through subspace-valued descriptors~\cite{grey2023separable, doronina2025, grey2026explainable}.

This geometric viewpoint has direct implications for predictive \gls{uq} on the Grassmann manifold by means of \gls{cp}.
In particular, prediction regions must be invariant to the particular orthonormal basis used to represent a response. Otherwise, discrepancies in coordinate representations will be mistakenly penalized as predictive errors, leading to artificially inflated prediction regions.
Consequently, \gls{cp} regions should respect the quotient geometry of the Grassmann manifold, characterizing uncertainty through changes in the principal directions that define the underlying manifold structure.
Furthermore, because the accuracy of surrogate models frequently varies across the input space, these geometric prediction regions must also be adaptive, scaling their size in response to local variations in predictive difficulty.
This work aims to contribute exactly in this direction by developing a framework for the predictive \gls{uq} of responses on the Grassmann manifold based on locally adaptive \gls{cp}.

\subsection{Related work}
\label{sec:intro-related}
Substantial work has focused on constructing surrogate models for high-dimensional outputs by first identifying informative low-dimensional representations by means of nonlinear dimension reduction and manifold learning. 
Nonlinear dimension-reduction techniques have been coupled with regression models to address high-dimensional parametric problems~\cite{kalogeris2020diffusion, lataniotis2020extending}. 
Developments in manifold learning specifically for \gls{uq} include reduced representations of high-dimensional black-box models~\cite{kontolati2022survey}, probabilistic learning on manifolds~\cite{soize2022probabilistic}, and diffusion-based probabilistic learning strategies for nonlinear manifolds~\cite{giovanis2026enabling}.  
These methods demonstrate the value of learning and propagating uncertainty on intrinsic low-dimensional representations, but they do not directly provide predictive \gls{uq} metrics for manifold-valued responses.

Grassmann manifolds have received particular attention when high-dimensional responses are naturally represented through subspaces. 
Grassmannian representations have been used to quantify variability in high-dimensional model responses~\cite{giovanis2018uncertainty}, construct Gaussian-process surrogate models directly from subspace-valued data~\cite{giovanis2020data}, and combine manifold learning with polynomial-chaos and geometric-harmonics surrogate models~\cite{kontolati2022manifold, giovanis2024polynomial, dos2022grassmannian_IJNM}. 
The Grassmannian also serves as a useful encoding of shape when representing planar curves~\cite{grey2023separable, grey2026explainable}.
These studies establish the Grassmann manifold as a natural representation space for reduced-order bases, shape representations, or other subspace-valued quantities of interest, without however providing predictive \gls{uq} metrics. 
Quantifying regression uncertainty directly on the Grassmannian has been pursued by means of Kalman filtering~\cite{hong2017regression} or Gaussian processes~\cite{zhang2021gaussian}.
In both these cases, predictive \gls{uq} is model-dependent, in contrast to the model-agnostic nature of \gls{cp}.

Focusing on \gls{cp} for regression models, inductive or split \gls{cp} offers a particularly practical framework for predictive \gls{uq}, which also provides finite-sample marginal coverage guarantees under data exchangeability~\cite{papadopoulos2002inductive}.
Additionally, normalized non-conformity scores have been introduced to accommodate input-dependent predictive difficulty through locally adaptive prediction intervals~\cite{papadopoulos2008normalized}. 
Subsequent work has further developed adaptive \gls{cp} for regression~\cite{lei2018distribution, wang2018fast, deutschmann2024adaptive} and extended it to multi-output settings~\cite{diquigiovanni2022conformal, dheur2025unified}. 
However, these works address either scalar or Euclidean-valued responses only.

More recently, \gls{cp} has extended to non-Euclidean and manifold-valued settings. 
Kuleshov et al.~\cite{kuleshov2018conformal} considered \gls{cp} in the context of manifold learning, using non-conformity scores based on the Bregman distance.
Rozenfeld and Goldshtein~\cite{rozenfeld2025conformal} developed \gls{cp} for manifold-based sound source localization using Gaussian processes. 
Cholaquidis et al.~\cite{cholaquidis2025conformal} studied \gls{cp} for regression on general Riemannian manifolds, providing a broader statistical treatment of manifold-valued responses.  
Shahbazi and Baheri~\cite{shahbazi2026geometry} introduced an adaptive geodesic \gls{cp} method for \gls{uq} on spherical geometries.
Last, Jaber et al.~\cite{jaber2026conformal} utilized \gls{cp} to quantify the model-form uncertainty of reduced-order models based on \gls{pod}. 
These developments demonstrate increasing interest in extending \gls{cp} to the manifold-learning setting.
The present work complements and extends these research directions, focusing specifically on locally adaptive \gls{cp} for Grassmann-valued responses. 
Our emphasis is on defining geometry-aware non-conformity scores tailored to the Grassmann manifold, but also on constructing input-dependent Grassmannian prediction regions whose size adapts to local prediction difficulty.
This combination, along with the specific contributions outlined next, distinguishes the present work from existing approaches in the literature.

\subsection{Contribution}
\label{sec:intro-contribution}
We formulate a locally adaptive \gls{cp} method that works directly on the Grassmann manifold.
To that end, we employ the split \gls{cp} variant~\cite{papadopoulos2002inductive} and define scalar non-conformity scores through basis-invariant geometric distances \cite{hamm2008grassmann}. 
The non-conformity scores are additionally scaled by an input-dependent difficulty estimator to produce prediction regions with sizes that adapt to local prediction difficulty.
The estimator is learned from log-transformed, geometric \gls{cv}  residuals, providing a strictly positive out-of-sample proxy for local prediction difficulty.
The resulting prediction regions are basis-invariant metric balls centered at the predicted subspace, therefore respecting the underlying geometric properties of the manifold. 
In addition, the finite-sample marginal coverage guarantee of split \gls{cp} is retained. 

The suggested framework is first validated on synthetic benchmarks designed to isolate geometric and input-dependent uncertainty. 
The results demonstrate that local adaptation preserves the prescribed marginal coverage while substantially improving empirical coverage uniformity across prediction-difficulty levels and, in strongly heteroscedastic settings, reducing the size of the resulting prediction regions.
Within these benchmarks, we additionally investigate the effect of the underlying Grassmannian distance metric. 
The methodology is further demonstrated on an engineering inverse design problem, where aerodynamic coefficients are mapped to Grassmann-valued airfoil representations. 
The application also illustrates how \gls{cp} regions constructed in the latent geometric space can be translated back into physically interpretable uncertainty envelopes, i.e., a range of plausible outcomes around the predicted airfoil shapes.

Note that the suggested framework is model-agnostic with respect to both the base Grassmann-valued regressor and the difficulty estimator. These components affect the adaptivity and efficiency of the prediction regions, but not the finite-sample marginal validity of split \gls{cp}.
Therefore, the methodology is not tied to the specific models used in the numerical investigations presented in this work.

\subsection{Paper organization}
The rest of this paper is organized as follows.
In Section~\ref{sec:split-cp}, we recall the basics of \gls{cp}, with emphasis on the split \gls{cp} variant of the method.
The necessary fundamentals on Grassmann manifolds and geometric \gls{ml} on them are presented in Section~\ref{sec:grassmann}.
Section~\ref{sec:acp-grassmann} presents the core methodological contribution of this work, namely, the suggested framework for \gls{cp} on Grassmann manifolds with locally adaptive prediction regions.
The method is first validated on synthetic benchmarks in Section~\ref{sec:validation}, and is subsequently applied to an airfoil inverse design problem in Section~\ref{sec:airfoil}.
The paper concludes with Section~\ref{sec:conclusion}, which summarizes its findings and suggests potential follow-up research. 

\section{Split conformal prediction}
\label{sec:split-cp}
In this section, we briefly review the standard split \gls{cp} framework for regression~\cite{papadopoulos2002inductive}. 
For exposition, standard split \gls{cp} is often introduced for a scalar response. 
Extensions to vector-valued responses are more complex and require dedicated approaches~\cite{diquigiovanni2022conformal, dheur2025unified}.
Moreover, this work is specifically concerned with structured responses residing on Grassmann manifolds, thus necessitating geometric adaptations in the definition of the non-conformity score and prediction region. 
These adaptations are detailed in Section~\ref{sec:acp-grassmann}. 
The standard formulation for scalar outputs is presented below as a mathematical baseline.

We denote with $\widehat{\mu}$ a regression model trained on a dataset $\mathcal{D}_{\mathrm{train}} = \left\{\left(\mathbf{x}^{(m)}, y^{(m)}\right)\right\}_{m=1}^M$, with input $\mathbf{x} \in \mathbb{R}^N$ and response $y \in \mathbb{R}$. 
The goal of \gls{cp} is to complement the model's prediction $\widehat{\mu}\left(\mathbf{x}^*\right)$ for a test input $\mathbf{x}^* \not\in \mathcal{D}_{\mathrm{train}}$, with a prediction interval $\widehat{C}_{1-\alpha}\left(\mathbf{x}^*\right)$ that contains the true response $y^*$ with a probability at least equal to $(1-\alpha)$.
Here, $\alpha \in \left(0,1\right)$ is a prescribed significance level.
Formally, the prediction interval must satisfy the condition
\begin{equation}
\label{eq:coverage-condition}
\mathbb{P}\left\{y^* \in \widehat{C}_{1-\alpha}\left(\mathbf{x}^*\right)\right\} \geq 1 - \alpha,
\end{equation}
where the probability is taken over the joint distribution of the data used to calibrate the prediction interval and the test point.

We focus on split \gls{cp}, which is preferred here due to its implementation simplicity and model training efficiency, compared to other \gls{cp} variants such as full \gls{cp}~\cite{gazin2024transductive} and cross \gls{cp}~\cite{vovk2015cross, vovk2018cross}.
Therein, a single regression model $\widehat{\mu}$ is trained on $\mathcal{D}_{\mathrm{train}}$ and held fixed afterwards.
A disjoint \emph{calibration} dataset $\mathcal{D}_{\mathrm{cal}} = \left\{\left(\mathbf{x}^{(j)}, y^{(j)}\right)\right\}_{j=1}^J$ is then used to compute so-called \emph{non‑conformity scores}~\cite{kato2023review}, which quantify the discrepancy between the true responses and the model's predictions.
The non-conformity measure is denoted with $s\left(\mathbf{x},y;\widehat{\mu}\right)$ and the non-conformity scores with $s_j = s\left(\mathbf{x}^{(j)}, y^{(j)};\widehat{\mu}\right)$, $j=1,\dots,J$. 
The $(1-\alpha)$-quantile of $\left(s_j\right)_{j=1}^J$ is denoted as $\widehat{q}_{1-\alpha}$ and set to the $k$-th smallest non-conformity score, where $k = \lceil (J+1)(1-\alpha) \rceil$. 
The prediction interval is then given by
\begin{equation}
\label{eq:split-cp-interval-general}
\widehat C_{1-\alpha}\left(\mathbf{x}^*\right) \;=\; \left\{\, y : \; s\left(\mathbf{x}^*,y;\widehat{\mu}\right) \le \widehat q_{1-\alpha}\,\right\}.
\end{equation}

In regression tasks, it is common to use the absolute residuals as non-conformity scores,
\begin{equation}
\label{eq:abs-residual}
\left|r_j\right| = \left| y^{(j)} - \widehat{\mu}\left(\mathbf{x}^{(j)}\right) \right|.
\end{equation}
In that case, a symmetric interval with constant width equal to $2\widehat{q}_{1-\alpha}$ is obtained:  
\begin{equation}
\label{eq:split-cp-interval}
\widehat{C}_{1-\alpha}(\mathbf{x}^*) =
\left[\,\widehat{\mu}\left(\mathbf{x}^*\right) - \widehat{q}_{1-\alpha}, \; 
\widehat{\mu}\left(\mathbf{x}^*\right) + \widehat{q}_{1-\alpha}\,\right].
\end{equation}

A downside of these constant intervals is that they do not take into account variations in the model's predictive accuracy across the input space. 
In cases where the model's predictive capability is input-dependent, it is desirable to have \emph{adaptive} intervals \cite{lei2018distribution, wang2018fast, deutschmann2024adaptive}, i.e., wider in regions of the input space with high prediction uncertainty and, accordingly, narrower in high-confidence input regions.
This is accomplished with the use of the normalized non-conformity measure
\begin{equation}
\label{eq:normalized-score}
s^{\mathrm{n}}(\mathbf{x}, y; \widehat{\mu}) =
\frac{|y - \widehat{\mu}(\mathbf{x})|}{\widehat{\eta}(\mathbf{x})},
\end{equation}
where the absolute residual is scaled by a so-called \emph{difficulty estimator} $\widehat{\eta}(\mathbf{x}) > 0$~\cite{papadopoulos2008normalized}.
This scaling results in intervals with adaptive width depending on the local prediction difficulty, given as
\begin{equation}
\label{eq:split-cp-interval-adaptive}
\widehat{C}_{1-\alpha}(\mathbf{x}^*) =
\left[\,\widehat{\mu}\left(\mathbf{x}^*\right) - \widehat{q}_{1-\alpha}^{\mathrm{n}} \widehat{\eta}\left(\mathbf{x}^*\right), \; 
\widehat{\mu}\left(\mathbf{x}^*\right) + \widehat{q}_{1-\alpha}^{\mathrm{n}} \widehat{\eta}\left(\mathbf{x}^*\right)\,\right],
\end{equation}
where the quantile $\widehat{q}_{1-\alpha}^{\mathrm{n}}$ is computed from the normalized non-conformity scores $\left(s_j^{\mathrm{n}}\right)_{j=1}^J$ using the same ordering rule as $\widehat{q}_{1-\alpha}$.

A key assumption of split \gls{cp} is the exchangeability of the calibration and test data, such that their joint distribution is invariant under sample permutations.
Under the exchangeability assumption, split \gls{cp} provides a finite‑sample marginal coverage guarantee at the target coverage level of $(1-\alpha)$. 
Note that the coverage guarantee is marginal and not conditional, meaning that the coverage condition \eqref{eq:coverage-condition} is satisfied on average over repeatedly constructed \gls{cp} intervals.
Exact conditional coverage cannot be achieved without strong distributional assumptions~\cite{vovk2012conditional, barber2021limits}. 

\section{Grassmann manifolds and geometric machine learning}
\label{sec:grassmann}
In this work, the responses of interest are not scalar values or standard Euclidean vectors, but rather structured features that reside on a Grassmann manifold~\cite{absil2004riemannian}, for example, dominant modes, subspaces, reduced-order bases, or objects encoding shape. 
In this section, we briefly review the necessary representations and metric properties of this space. 

\subsection{Representation and principal angles}
\label{sec:grassmann-representation}
The Grassmann manifold, also referred to as the Grassmannian and denoted $\mathrm{Gr}(p, n)$, is defined as the set of all $p$-dimensional linear subspaces within an $n$-dimensional vector space $\mathbb{R}^n$ (where $0 < p < n$). 
A point (i.e., subspace) $\mathcal{Y} \in \mathrm{Gr}(p, n)$ is typically represented by an orthonormal matrix $\mathbf{Y} \in \mathbb{R}^{n \times p}$ spanning the subspace, such that $\mathbf{Y}^\top \mathbf{Y} = \mathbf{I}_p$, where $\mathbf{I}_p$ is the $p \times p$ identity matrix. 
A full-rank basis matrix representing a subspace is not unique: right multiplication by any invertible matrix in the general linear group of degree $p$, $\mathrm{GL}(p)$, produces another basis spanning the same subspace. 
After restricting to orthonormal representatives, this non-uniqueness reduces to the right action of the orthogonal group of degree $p$, $\mathrm{O}(p)$~\cite{absil2004riemannian}. 
For example, two orthonormal bases, $\mathbf{Y}_1, \mathbf{Y}_2$, represent the same point in $\mathrm{Gr}(p, n)$ if there exists an orthogonal matrix $\mathbf{Q}$ such that $\mathbf{Y}_1 = \mathbf{Y}_2 \mathbf{Q}$. 
Alternatively, $\mathcal{Y}$ can be uniquely represented by its $n \times n$ orthogonal projection matrix $\mathbf{P}_{\mathcal{Y}} = \mathbf{Y} \mathbf{Y}^\top$, which is basis-invariant $\left( (\mathbf{Y}\mathbf{Q})(\mathbf{Y}\mathbf{Q})^{\top} = \mathbf{Y}\mathbf{Q}\mathbf{Q}^{\top}\mathbf{Y}^{\top} = \mathbf{Y}\mathbf{Y}^{\top} = \mathbf{P}_{\mathcal{Y}} \right)$, symmetric $\left(\mathbf{P}_{\mathcal{Y}}^\top = \mathbf{P}_{\mathcal{Y}} \right)$, idempotent $\left(\mathbf{P}_{\mathcal{Y}}^2 = \mathbf{P}_{\mathcal{Y}}\right)$, and has rank $p$ $\left(\mathrm{tr}(\mathbf{Y}\mathbf{Y}^{\top}) = \mathrm{tr}(\mathbf{Y}^{\top} \mathbf{Y}) = \mathrm{tr}(\mathbf{I}_p) = p\right)$. 

Geometrically, the distance between any pair of subspaces $\mathcal{Y}_1, \mathcal{Y}_2 \in \mathrm{Gr}(p,n)$ is characterized by their \emph{principal angles} $\theta_1, \dots, \theta_p \in [0, \pi/2]$. 
Given their respective (representative) orthonormal bases $\mathbf{Y}_1$ and $\mathbf{Y}_2$, the principal angles are computed via the \gls{svd},
\begin{equation}
\label{eq:principal-angles}
\mathbf{Y}_1^\top \mathbf{Y}_2 = \mathbf{U} \mathbf{\Sigma} \mathbf{V}^\top,
\end{equation}
where $\mathbf{\Sigma} = \mathrm{diag}(\sigma_1, \dots, \sigma_p)$ contains the singular values, and $\mathbf{U}, \mathbf{V}$ are orthogonal matrices. 
The principal angles are defined as $\theta_i = \arccos(\sigma_i)$, $i = 1, \dots, p$.

\subsection{Grassmannian distance metrics}
\label{sec:grassmann-metrics}
Because $\mathrm{Gr}(p,n)$ is a curved Riemannian manifold, the standard Euclidean distance between arbitrary basis representatives is not well defined as a distance between Grassmann points because it depends on the selected bases. 
Instead, distances are defined as functions of the principal angles. 
In this work, we consider four Grassmannian distance metrics~\cite{hamm2008grassmann}, namely:
\begin{subequations}
\label{eq:grassmann-distances}
\begin{align}
\text{Geodesic distance:}&  &&d_{\mathrm{geo}} = \|\boldsymbol{\theta}\|_2 = \sqrt{\sum_{i=1}^p \theta_i^2} \\ 
\text{Projection metric:}&  &&d_{\mathrm{proj}} = \frac{1}{\sqrt{2}} \| \mathbf{P}_{\mathcal{Y}_1} - \mathbf{P}_{\mathcal{Y}_2} \|_{\mathrm{F}} = \sqrt{\sum_{i=1}^p \sin^2(\theta_i)} \\
\text{Procrustes metric:}&  &&d_{\mathrm{Proc}} = \min_{\substack{\mathbf{Q} \in \mathbb{R}^{p \times p} \\ \mathbf{Q}^\top\mathbf{Q} = \mathbf{I}_p}}\| \mathbf{Y}_1 - \mathbf{Y}_2 \mathbf{Q} \|_{\mathrm{F}} = \sqrt{2 \sum_{i=1}^p (1 - \cos\theta_i)} \\
\text{Binet-Cauchy metric:}&  &&d_{\mathrm{BC}} = \sqrt{1 - \prod_{i=1}^p \cos^2\theta_i} 
\end{align}
\end{subequations}
where $\| \cdot \|_{\mathrm{F}}$ denotes the Frobenius norm and  $\mathbf{Y}_1, \mathbf{Y}_2 \in \mathbb{R}^{n \times p}$ are (representative) orthonormal bases spanning the subspaces $\mathcal{Y}_1, \mathcal{Y}_2$.
Each metric captures different geometric properties: $d_{\mathrm{geo}}$ is defined as the length of the shortest curve connecting the two subspaces; $d_{\mathrm{proj}}$ measures an embedded Euclidean deviation between the orthogonal projection operators of the subspaces; $d_{\mathrm{Proc}}$ measures the minimal Euclidean distance between the bases after optimal orthogonal alignment; and $d_{\mathrm{BC}}$ is a volume-based discrepancy metric. 

Among these metrics, \(d_{\rm geo}\) is the intrinsic Riemannian distance associated with the canonical Grassmann metric~\cite{absil2004riemannian}.
Also note that, despite the Euclidean nature of the embedding space for $d_{\mathrm{proj}}$, the space of projection matrices is not closed under linear combinations because of the idempotent property. 
Instead, the ability to isometrically embed into a Euclidean space with the projection operators motivates a distinct notion of ``shortest'' path which ``cuts the corner'' much like the chord of a circle. 
For this reason, $d_{\mathrm{proj}}$ is often referred to as the chordal distance.

\subsection{Regression on the Grassmann manifold}
\label{sec:grassmann-regression}
In supervised manifold learning, given a dataset $\mathcal{D}_{\mathrm{train}} = \{(\mathbf{x}^{(m)}, \mathbf{Y}^{(m)})\}_{m=1}^M$ where $\mathbf{x} \in \mathbb{R}^N$ and $\mathbf{Y} \in \mathbb{R}^{n \times p}$ spans the subspace $\mathcal{Y} \in \mathrm{Gr}(p,n)$, the goal is to learn a regression map $\widehat{\mu}: \mathbb{R}^N \to \mathbb{R}^{n \times p}$ that maps input features to a representative, orthonormal basis. 
Because linear combinations of orthogonal matrices are not closed, standard Euclidean averaging is not helpful. 
Instead, predictions with orthonormal constraints (Stiefel) and subsequent orthogonal invariance as rotations and reflections through right actions on the basis (Grassmannian) must be formulated geometrically~\cite{edelman1998geometry}. 

A broad class of regression models, ranging from local smoothers to deep neural networks, can be adapted to the manifold setting by formulating the prediction as a conditional \emph{Fréchet mean} \cite{dubey2019frechet}.
Given a metric $d$ on $\mathrm{Gr}(p,n)$, a conditional Fréchet estimator may be defined as
\begin{equation}
\widehat{\mathcal{Y}}(\mathbf{x}) \in \argmin_{\mathcal{Y}\in\mathrm{Gr}(p,n)} \sum_{m=1}^{M} w_m(\mathbf{x})\, d^2\!\left(\mathcal{Y},\mathcal{Y}^{(m)}\right), 
\end{equation}
where $w_m(\mathbf{x}) \geq 0$ are input-dependent weights, typically normalized so that $\sum_{m=1}^{M}w_m(\mathbf{x})=1$. 
When $d=d_{\mathrm{geo}}$, this corresponds to the intrinsic Fréchet objective associated with the canonical Riemannian geometry of the Grassmann manifold. 
Other basis-invariant Grassmannian metrics may instead define alternative, including extrinsic, notions of centrality. 
Using orthonormal representative bases $\mathbf{Y}, \mathbf{Y}^{(m)} \in \mathbb{R}^{n\times p}$, the same estimator can be written as
\begin{equation}
\label{eq:frechet-mean}
\widehat{\mu}(\mathbf{x}) \in \argmin_{\substack{\mathbf{Y} \in \mathbb{R}^{n \times p} \\ \mathbf{Y}^\top\mathbf{Y} = \mathbf{I}_p}} \sum_{m=1}^M w_m(\mathbf{x}) \, d_{\mathrm{Gr}}^2\left(\mathbf{Y}, \mathbf{Y}^{(m)}\right),
\end{equation}
where $d_{\mathrm{Gr}}$ is a basis-invariant Grassmannian distance metric, so that the objective depends only on the corresponding subspaces rather than on the particular orthonormal bases.

The theoretical global minimizer of \eqref{eq:frechet-mean} is known as the \emph{Fréchet mean}. 
However, because closed-form solutions do not generally exist for curved spaces, one typically solves for a local minimizer known as the \emph{Karcher mean}. 
Finding the Karcher mean requires Riemannian gradient descent \cite{gao2022learning}, which replaces standard Euclidean addition and subtraction with two fundamental geometric operations: the \emph{logarithmic map} and the \emph{exponential map}. 
Let us denote the tangent space to $\mathrm{Gr}(p,n)$ evaluated  at the base point $\mathcal{Y} \in \mathrm{Gr}(p,n)$ as $\mathcal{T}_{\mathcal{Y}} \mathrm{Gr}(p,n)$. 
Given two elements $\mathcal{Y}_1, \mathcal{Y}_2 \in \mathrm{Gr}(p,n)$, where $\mathcal{Y}_1$ serves as the base point, the logarithmic map returns a tangent vector $\mathrm{Log}_{\mathcal{Y}_1}(\mathcal{Y}_2) \in \mathcal{T}_{\mathcal{Y}_1}\mathrm{Gr}(p,n)$, whose norm matches the geodesic distance between $\mathcal{Y}_1$ and $\mathcal{Y}_2$. 
Accordingly, the exponential map projects a tangent vector forward along geodesics of the Grassmann manifold, yielding $\mathrm{Exp}_{\mathcal{Y}_1}(\Delta) \in \mathrm{Gr}(p,n)$ for some $\Delta \in \mathcal{T}_{\mathcal{Y}_1}\mathrm{Gr}(p,n)$. 
Within an injectivity radius of $\pi/2$, the exponential map is a diffeomorphism from the open ball of radius $\pi/2$ centered at the origin on $\mathcal{T}_{\mathcal{Y}_1}\mathrm{Gr}(p,n)$ onto the open geodesic ball of radius $\pi/2$ centered at $\mathcal{Y}_1$ on $\mathrm{Gr}(p,n)$.

Algorithmically, the Karcher mean iterates from an initial guess of central tangent space at a mean as follows: (i) compute the arithmetic (Euclidean) average over the logarithmic map applied to each data point; (ii) step in the average direction with the exponential map to update the estimated mean; (iii) repeat until satisfying a convergence criterion. 
Grey et al.~\cite{grey2023separable} provide an example application, while elaborations on the theoretical concepts can be found in standard texts~\cite{absil2004riemannian, edelman1998geometry}. 
Computing this intrinsic Karcher mean requires iterative SVD decompositions for every Log/Exp evaluation, making it potentially computationally expensive.
As a computationally cheaper alternative, one may use an extrinsic Fréchet estimator induced by the projection embedding.
By mapping each target subspace to its unique projection matrix $\mathbf{P}^{(m)} = \mathbf{Y}^{(m)} (\mathbf{Y}^{(m)})^\top$, the conditional extrinsic mean is computed in a single step:
\begin{equation}
\label{eq:extrinsic-mean}
\bar{\mathbf{P}}(\mathbf{x}) = \sum_{m=1}^M w_m(\mathbf{x}) \mathbf{P}^{(m)}.
\end{equation}
Because the arithmetic average $\bar{\mathbf{P}}(\mathbf{x})$ is generally not a valid rank-$p$ orthogonal projector, the final predicted subspace basis $\widehat{\mathbf{Y}} = \widehat{\mu}(\mathbf{x})$ is obtained by extracting the eigenvectors corresponding to the $p$ largest eigenvalues of $\bar{\mathbf{P}}(\mathbf{x})$. However, intrinsic and extrinsic means generally represent distinct notions of centrality~\cite{srivastava2002monte,fletcher2004principal}. 
Regardless, the extrinsic projection mean serves as a fast, computationally tractable geometric estimator that can be easily coupled with any weight-generating learning algorithm.

\section{Locally adaptive conformal prediction on Grassmann manifolds}
\label{sec:acp-grassmann}
While the generalized geometric \gls{ml} regression framework described in Section~\ref{sec:grassmann} provides an estimate for an orthonormal basis $\widehat{\mathbf{Y}} = \widehat{\mu}(\mathbf{x}) \in \mathbb{R}^{n \times p}$, it lacks a native mechanism to quantify prediction uncertainty. 
To address this, we introduce an adaptive split \gls{cp} framework tailored to Grassmann-valued outputs.

\subsection{Grassmannian non-conformity scores and conformal regions}
\label{sec:grassmann-scores-regions}
First, the geometric base regressor  is trained on $\mathcal{D}_{\mathrm{train}}$ to map an input vector to a representative orthonormal basis matrix of the predicted subspace, yielding the mapping
\begin{equation}
\mathbf{x} \: \longmapsto \: \widehat{\mu}(\mathbf{x}) \: \longmapsto \: \widehat{\mathbf{Y}} \in \mathbb{R}^{n \times p} \: \longmapsto \: \widehat{\mathcal{Y}} \in \mathrm{Gr}(p,n).
\end{equation}
Given a test input $\mathbf{x}^* \not \in \mathcal{D}_{\mathrm{train}}$, we seek a prediction region $\widehat{C}_{1-\alpha}(\mathbf{x}^*) \subset \mathrm{Gr}(p,n)$ that satisfies the marginal coverage condition
\begin{equation}
\label{eq:coverage-condition-Gr}
\mathbb{P}\left\{ \mathcal{Y}^* \in \widehat{C}_{1-\alpha}\left(\mathbf{x}^*\right) \right\} \geq 1 - \alpha.
\end{equation}

Following the split \gls{cp} framework introduced in Section~\ref{sec:split-cp}, we use the disjoint calibration dataset $\mathcal{D}_{\mathrm{cal}}$ to compute scalar non-conformity scores $s_j$, $j = 1, \dots, J$. 
These scores are defined via a Grassmannian distance between the target and predicted basis matrices as
\begin{equation}
\label{eq:grassmann-score}
s_j = d_{\mathrm{Gr}} \left( \mathbf{Y}^{(j)}, \widehat{\mu}\left(\mathbf{x}^{(j)}\right) \right),
\end{equation}
where $d_{\mathrm{Gr}}$ may correspond to any of the metrics introduced in Section~\ref{sec:grassmann-metrics}.
The conformal quantile $\widehat{q}_{1-\alpha}$ can be computed from $\left(s_j\right)_{j=1}^J$ using the same empirical ordering rule as in Section~\ref{sec:split-cp}. 
Then, the standard, non-adaptive prediction region can be defined as a metric ball $\mathcal{B}\left(\widehat{\mathcal{Y}}^*, R \right)$ with center $\widehat{\mathcal{Y}}^* = \mathrm{span}\left( \widehat{\mu}\left(\mathbf{x}^*\right) \right)$ and a constant radius $R = \widehat{q}_{1-\alpha}$:
\begin{equation}
\label{eq:standard-ball-set}
\widehat{C}_{1-\alpha}\left(\mathbf{x}^*\right) = \mathcal{B}\left(\widehat{\mathcal{Y}}^*, \widehat{q}_{1-\alpha}\right) = \left\{ \mathcal{Y} \in \mathrm{Gr}(p,n) : d_{\mathrm{Gr}}\left(\mathbf{Y}, \widehat{\mu}(\mathbf{x}^*)\right) \le \widehat{q}_{1-\alpha} \right\},
\end{equation}
where $\mathbf{Y} \in \mathbb{R}^{n \times p}$ is any orthonormal basis matrix representing the subspace $\mathcal{Y}$.

Similar to the standard split \gls{cp} setting described in Section~\ref{sec:split-cp}, the constant radius of the prediction region \eqref{eq:standard-ball-set} does not account for settings where the geometric dispersion of the target subspace $\mathcal{Y}^*$ varies conditionally on $\mathbf{x}^*$.
To adaptively scale the prediction regions similarly to \eqref{eq:split-cp-interval-adaptive}, a (strictly positive) difficulty estimator function $\widehat{\eta}$ must be used to compute normalized non-conformity scores
\begin{equation}
\label{eq:grassmann-score-normalized}
s_j^{\mathrm{n}} = \frac{d_{\mathrm{Gr}} \left( \mathbf{Y}^{(j)}, \widehat{\mu}\left(\mathbf{x}^{(j)}\right) \right)}{\widehat{\eta}\left(\mathbf{x}^{(j)}\right)}.
\end{equation}
Accordingly, the quantile $\widehat{q}_{1-\alpha}^{\mathrm{n}}$ is computed from $\left(s_j^{\mathrm{n}}\right)_{j=1}^J$ and
locally adaptive prediction regions are obtained, corresponding to metric balls $\mathcal{B}\left(\widehat{\mathcal{Y}}^*, R^* \right)$ with input-dependent radii $R^* = \widehat{q}_{1-\alpha}^{\mathrm{n}} \widehat{\eta} \left( \mathbf{x}^*\right)$: 
\begin{equation}
\label{eq:adaptive-ball-set}
\widehat{C}_{1-\alpha}\left(\mathbf{x}^*\right) = \mathcal{B}\left(\widehat{\mathcal{Y}}^*, \widehat{q}_{1-\alpha}^{\mathrm{n}} \, \widehat{\eta}\left(\mathbf{x}^*\right)\right) = \left\{ \mathcal{Y} \in \mathrm{Gr}(p,n) : d_{\mathrm{Gr}}\left(\mathbf{Y}, \widehat{\mu}(\mathbf{x}^*)\right) \le \widehat{q}_{1-\alpha}^{\mathrm{n}} \widehat{\eta} \left( \mathbf{x}^*\right)\right\}.
\end{equation}
Section~\ref{sec:difficulty-estimator} is dedicated to computing the difficulty estimator $\widehat{\eta}$.

\subsection{Difficulty estimator computation}
\label{sec:difficulty-estimator}
To construct the locally adaptive \gls{cp} regions defined in \eqref{eq:adaptive-ball-set}, a difficulty estimator function $\widehat{\eta}: \mathbb{R}^N \to \mathbb{R}^+$ is required, which provides a proxy for the local scale of the geometric prediction error of the base regression model $\widehat{\mu}$ at a given input $\mathbf{x}$.
A naive approach would be to train $\widehat{\eta}$ directly on the Grassmannian training residuals (i.e., distances) 
\begin{equation}
\label{eq:train-residuals-grassmann}
e_m = d_{\mathrm{Gr}}\left(\mathbf{Y}^{(m)}, \widehat{\mu}\left(\mathbf{x}^{(m)}\right)\right). 
\end{equation}
However, as regression models tend to overfit on their training data, this approach would lead to systematic underestimation of out-of-sample uncertainty. 
Additionally, as Grassmannian distances may be concentrated near zero, training a model directly on the raw residuals can result in numerical instabilities or yield invalid negative predictions.

To obtain an out-of-sample proxy for the local prediction error, a $K$-fold \gls{cv} procedure is followed~\cite{shahbazi2026geometry}.
First, $\mathcal{D}_{\mathrm{train}}$ is randomly partitioned into $K$ disjoint subsets (folds) $\mathcal{D}_k \subset \mathcal{D}_{\mathrm{train}}$, $k=1,\dots,K$, such that $\mathcal{D}_{\mathrm{train}} = \bigcup_{k=1}^K \mathcal{D}_k$ and $\mathcal{D}_k \cap \mathcal{D}_l = \varnothing$ for $k \neq l$.
For each fold $k \in \{1, \dots, K\}$, a fold-out base model $\widehat{\mu}_{-k}$ is trained on the fold-out dataset $\mathcal{D}_{-k} = \mathcal{D}_{\mathrm{train}} \setminus \mathcal{D}_k$.
Then, for each sample $\left(\mathbf{x}^{(m)}, \mathbf{Y}^{(m)}\right) \in \mathcal{D}_{k}$, the Grassmannian \gls{cv} residual is computed as 
\begin{equation}
\label{eq:cv-residual-grassmann}
e_m^{\mathrm{CV}} =  d_{\mathrm{Gr}}\left(\mathbf{Y}^{(m)}, \widehat{\mu}_{-k}\left(\mathbf{x}^{(m)}\right)\right).
\end{equation}
In the numerical experiments presented in Sections~\ref{sec:validation} and \ref{sec:airfoil}, we consistently set $K=5$.

Subsequently, the \gls{cv} residuals are log-transformed as
\begin{equation}
\label{eq:cv-log-residual-grassmann}
z_m = \ln \left( e_m^{\mathrm{CV}} + \delta \right),
\end{equation}
where $\delta > 0$ is a numerical safeguard against vanishing residuals.
The new dataset composed of strictly out-of-fold scalar log-residuals is denoted as $\mathcal{D}_z = \left\{\left(\mathbf{x}^{(m)}, z_m\right)\right\}_{m=1}^M$. 
An auxiliary scalar regression model $\widehat{\gamma}: \mathbb{R}^N \to \mathbb{R}$ is then trained on $\mathcal{D}_z$ to map the input features to the expected logarithmic geometric error. 
Given a new input vector $\mathbf{x}$, the final strictly positive difficulty estimator is obtained by mapping the predicted log-transformed error back to the native scale via the exponential function, such that 
\begin{equation}
\label{eq:final-difficulty-estimator}
\widehat{\eta}(\mathbf{x}) = \exp\left( \widehat{\gamma}(\mathbf{x}) \right).
\end{equation}
The positivity requirement of the normalized non-conformity scores in \eqref{eq:grassmann-score-normalized} is then naturally satisfied.

\subsection{Marginal coverage on the Grassmannian}
\label{sec:marginal-coverage-grassmann}
The primary advantage of the split \gls{cp} framework is that it provides finite-sample, distribution-free coverage guarantees without requiring any assumptions about the underlying data generating process, the convergence of the base model $\widehat{\mu}$, or the accuracy of the difficulty estimator $\widehat{\eta}$. 
The validity of the adaptive prediction region relies entirely on the exchangeability of the calibration and test data.

Let the calibration dataset $\mathcal{D}_{\mathrm{cal}} = \{(\mathbf{x}^{(j)}, \mathbf{Y}^{(j)})\}_{j=1}^J$ and the test point $(\mathbf{x}^*, \mathbf{Y}^*)$ be exchangeable. 
Define the normalized non-conformity scores $s_j^{\mathrm{n}}$ as in \eqref{eq:grassmann-score-normalized} for all $j=1, \dots, J$, and let $\widehat{q}_{1-\alpha}^{\mathrm{n}}$ be the $k$-th smallest value among these scores, where $k = \lceil(J+1)(1-\alpha)\rceil$. 
When \(k\le J\), let \(\widehat q_{1-\alpha}^{\mathrm n}\) be the \(k\)-th smallest calibration score; when \(k=J+1\), set  $\widehat{q}_{1-\alpha}^{\mathrm{n}}=+\infty$.
Then, the adaptive \gls{cp} region $\widehat{C}_{1-\alpha}(\mathbf{x}^*)$ defined in \eqref{eq:adaptive-ball-set} guarantees the target (marginal) coverage:
\begin{equation}
\label{eq:marginal-coverage-guarantee-grassmann}
\mathbb{P}\left\{ \mathcal{Y}^* \in \widehat{C}_{1-\alpha}(\mathbf{x}^*) \right\} \geq 1 - \alpha.
\end{equation}
Furthermore, if the normalized non-conformity scores are almost surely distinct, the coverage is bounded above by $1 - \alpha + \frac{1}{J+1}$.

By the design of the split \gls{cp} framework, the base regressor $\widehat{\mu}$ and the difficulty estimator $\widehat{\eta}$ are fitted strictly on the disjoint training set $\mathcal{D}_{\mathrm{train}}$. 
Consequently, when conditioning on $\mathcal{D}_{\mathrm{train}}$, the functions $\widehat{\mu}$ and $\widehat{\eta}$ are deterministic. 
Because the $J$ calibration points and the test point are exchangeable, the $J+1$ scalar non-conformity scores $\{s_1^{\mathrm{n}}, \dots, s_J^{\mathrm{n}}, s_*^{\mathrm{n}}\}$ are conditionally exchangeable real-valued random variables. 
The event $\mathcal{Y}^* \in \widehat{C}_{1-\alpha}(\mathbf{x}^*)$ is algebraically equivalent to the event $s_*^{\mathrm{n}} \le \widehat{q}_{1-\alpha}^{\mathrm{n}}$. 
By the standard properties of exchangeable random variables, the probability that the test score $s_*^{\mathrm{n}}$ is among the $\lceil(J+1)(1-\alpha)\rceil$ smallest values of the $J+1$ scores is bounded below by $1-\alpha$ \cite{vovk2005algorithmic}.

\paragraph{Remark on marginal versus conditional coverage:} 
The coverage guarantee in \eqref{eq:marginal-coverage-guarantee-grassmann} is a \emph{marginal} one, meaning that the probability is taken over the joint distribution of $(\mathbf{x}, \mathcal{Y})$ as well as the random split of the data. 
It is well-established that exact, finite-sample \emph{conditional} coverage cannot be achieved without strict assumptions on the data distribution~\cite{barber2021limits}.
On the Grassmann manifold, conditional coverage would be defined as $\mathbb{P}\{ \mathcal{Y}^* \in \widehat{C}_{1-\alpha}(\mathbf{x}^*) \mid \mathbf{x}^* = \mathbf{x} \} \ge 1 - \alpha$, $\forall \mathbf{x}$.
By introducing the local difficulty estimator $\widehat{\eta}$ into the non-conformity score, the adaptive \gls{cp} framework aims to produce prediction regions that reflect local geometric uncertainty, often leading to improved conditional coverage in practice.

\paragraph{Remark on geodesic uniqueness:} The coverage guarantee in \eqref{eq:marginal-coverage-guarantee-grassmann} does not depend on geodesic uniqueness; it requires only that the chosen non-conformity score be a well-defined measurable function. 
For the canonical Grassmann geodesic distance, minimizing geodesics are unique only within the injectivity radius $\pi/2$. Beyond the cut locus, a minimizing geodesic need not be unique, although the geodesic distance remains well defined \cite{lee2018introduction}. 
Since the Grassmann manifold is compact and geodesically complete, at least one minimizing geodesic connects any two subspaces.

\subsection{Algorithmic summary}
\label{sec:algorithmic-summary}
We conclude the methodological framework by summarizing the complete, end-to-end procedure for adaptive \gls{cp} on Grassmann manifolds. Algorithm~\ref{alg:adaptive_grassmann_cp} details the sequential execution of the training, calibration, and prediction phases.
To emphasize the model-agnostic nature of the suggested framework, the abstract notation $\mathcal{A}_{\widehat{\mu}}$ and $\mathcal{A}_{\widehat{\gamma}}$ is used for the regression algorithms employed to train the base geometric regression model $\widehat{\mu}$ and the auxiliary regression model $\widehat{\gamma}$ employed within the difficulty estimator, respectively. 

\begin{algorithm}[t!]
    \caption{Adaptive Conformal Prediction on the Grassmannian}
    \label{alg:adaptive_grassmann_cp}
    \textbf{Inputs:}
    Training dataset $\mathcal{D}_{\mathrm{train}} = \left\{\left(\mathbf{x}^{(m)}, \mathbf{Y}^{(m)}\right)\right\}_{m=1}^M$ and calibration dataset $\mathcal{D}_{\mathrm{cal}} = \left\{\left(\mathbf{x}^{(j)}, \mathbf{Y}^{(j)}\right)\right\}_{j=1}^J$ ,
    test covariate $\mathbf{x}^*$,
    significance level $\alpha \in (0, 1)$,
    Grassmann distance $d_{\mathrm{Gr}}$ from \eqref{eq:grassmann-distances},
    base regressor algorithm $\mathcal{A}_{\widehat{\mu}}$ (see Section~\ref{sec:grassmann-regression}),
    auxiliary regressor algorithm $\mathcal{A}_{\widehat{\gamma}}$, number of folds $K$, and safeguarding constant $\delta > 0$ in \eqref{eq:cv-log-residual-grassmann}.

    \noindent\textbf{Output:}
    Prediction region $\widehat{C}_{1-\alpha}(\mathbf{x}^*) \subset \mathrm{Gr}(p,n)$ satisfying the marginal coverage guarantee \eqref{eq:marginal-coverage-guarantee-grassmann}.

    \begin{algorithmic}[1]
        \vspace{1mm}
        \Statex \textbf{\textit{Phase 1: Base model \& difficulty estimator training}} \hfill (Sections~\ref{sec:grassmann-regression}, \ref{sec:difficulty-estimator})
        \State Train the base model on the full training set: $\widehat{\mu} \leftarrow \mathcal{A}_{\widehat{\mu}}\left(\mathcal{D}_{\mathrm{train}}\right)$
        \State Partition $\mathcal{D}_{\mathrm{train}}$ into $K$ disjoint folds $\mathcal{D}_1, \dots, \mathcal{D}_K$
        \For{$k = 1, \dots, K$}
        \State Train fold-out model: $\widehat{\mu}_{-k} \leftarrow \mathcal{A}_{\widehat{\mu}}\left(\mathcal{D}_{\mathrm{train}} \setminus \mathcal{D}_k\right)$
        \For{each $(\mathbf{x}^{(m)}, \mathbf{Y}^{(m)}) \in \mathcal{D}_k$}
        \State CV residual: $e_m^{\mathrm{CV}} = d_{\mathrm{Gr}}\big(\mathbf{Y}^{(m)}, \widehat{\mu}_{-k}(\mathbf{x}^{(m)})\big)$ \Comment{Eq.~\eqref{eq:cv-residual-grassmann}}
        \State Log-transformed CV residual: $z_m = \ln\left(e_m^{\mathrm{CV}} + \delta\right)$ \Comment{Eq.~\eqref{eq:cv-log-residual-grassmann}}
        \EndFor
        \EndFor
        \State Train auxiliary regressor: $\widehat{\gamma} \leftarrow \mathcal{A}_{\widehat{\gamma}}\left(\left\{\left(\mathbf{x}^{(m)}, z_m\right)\right\}_{m=1}^M\right)$ \Comment{Section~\ref{sec:difficulty-estimator}}
        \State Difficulty estimator: $\widehat{\eta}\left(\mathbf{x}\right) = \exp\left(\widehat{\gamma}\left(\mathbf{x}\right)\right)$ \Comment{Eq.~\eqref{eq:final-difficulty-estimator}}

        \vspace{1mm}
        \Statex \textbf{\textit{Phase 2: Conformal calibration}} \hfill (Sections~\ref{sec:grassmann-scores-regions}, \ref{sec:marginal-coverage-grassmann})
        \For{$j = 1, \dots, J$}
        \State Predict basis: $\widehat{\mathbf{Y}}^{(j)} = \widehat{\mu}\left(\mathbf{x}^{(j)}\right)$
        \State Normalized score: $s_j^{\mathrm{n}} = d_{\mathrm{Gr}}\big(\mathbf{Y}^{(j)}, \widehat{\mathbf{Y}}^{(j)}\big) / \,\widehat{\eta}\left(\mathbf{x}^{(j)}\right)$ \Comment{Eq.~\eqref{eq:grassmann-score-normalized}}
        \EndFor
        \State Sort scores in ascending order: $s_{(1)}^{\mathrm{n}} \le s_{(2)}^{\mathrm{n}} \le \dots \le s_{(J)}^{\mathrm{n}}$
        \State Set $k_\alpha = \lceil (J+1)(1-\alpha) \rceil$ and $\widehat{q}_{1-\alpha}^{\mathrm{n}} \leftarrow s_{(k_\alpha)}^{\mathrm{n}}$ if $k_\alpha \le J$, else $\widehat{q}_{1-\alpha}^{\mathrm{n}} \leftarrow +\infty$ \Comment{Section~\ref{sec:marginal-coverage-grassmann}}

        \vspace{1mm}
        \Statex \textbf{\textit{Phase 3: Locally adaptive conformal prediction}} \hfill (Section~\ref{sec:grassmann-scores-regions})
        \State Predicted subspace: $\widehat{\mathcal{Y}}^* = \mathrm{span}\left(\widehat{\mu}(\mathbf{x}^*)\right)$
        \State Locally scaled radius: $R^* = \widehat{q}_{1-\alpha}^{\mathrm{n}} \, \widehat{\eta}(\mathbf{x}^*)$
        \State \Return $\widehat{C}_{1-\alpha}(\mathbf{x}^*) = \mathcal{B}\left(\widehat{\mathcal{Y}}^*, R^*\right)$ \Comment{Eq.~\eqref{eq:adaptive-ball-set}}
    \end{algorithmic}
\end{algorithm}

\section{Validation benchmarks}
\label{sec:validation}
To evaluate the proposed locally adaptive Grassmannian \gls{cp} framework, we first test it on two synthetic test-cases. 
We compare adaptive versus non-adaptive \gls{cp} regions for different distance metrics in terms of coverage and statistical efficiency. 
All metrics are computed on a held‑out test set $\mathcal{D}_{\mathrm{test}} = \{(\mathbf{x}^{(l)}, \mathbf{Y}^{(l)})\}_{l=1}^L$ that is disjoint from both $\mathcal{D}_{\mathrm{train}}$ and $\mathcal{D}_{\mathrm{cal}}$.
Before detailing the test-cases in Sections~\ref{sec:sphere} and \ref{sec:rot-plane}, respectively, we first define our evaluation metrics in Section~\ref{sec:eval-metrics}.

\subsection{Evaluation metrics}
\label{sec:eval-metrics}
\subsubsection{Conditional coverage estimation}
\label{sec:stratified-coverage-estimation}
In addition to (marginal) coverage estimation, we also estimate coverage uniformity, that is, how much the coverage probability varies across different levels of prediction difficulty. 
This is important because marginal coverage alone can hide serious conditional miscoverage. 
For example, a \gls{cp} method could achieve the target coverage by over‑covering easy-to-predict responses (coverage near $1$) and under‑covering hard-to-predict response (coverage far below the $1-\alpha$ target level).
This is undesirable, especially since reliable uncertainty estimates are primarily needed for difficult predictions.

Because exact point-wise conditional coverage is impossible without strong distributional assumptions~\cite{vovk2012conditional}, we rely on a heuristic approximation.
In particular, the test points are grouped into equal-frequency bins according to true or estimated difficulty, depending on whether the former is available. 
Then, the empirical coverage per bin and the standard deviation of coverage across bins are computed.
A low standard deviation signifies that the method provides uniform coverage across bins, meaning that easy and hard inputs are covered at a similar rate. 
This is the desired behavior of an adaptive \gls{cp} method.
Conversely, a high standard deviation indicates that coverage depends strongly on difficulty, which is typical of non‑adaptive \gls{cp} methods.
Using these heuristics, we show that, while our method strictly guarantees marginal coverage only, it empirically achieves strong approximate conditional coverage across the feature space.

\subsubsection{Cross-metric standardization and efficiency evaluation}
\label{sec:cross-metric-standardization}
When comparing adaptive and non-adaptive \gls{cp} methods, one often examines the size of the prediction regions. 
However, comparing relevant quantities such as coverage radius or area across different Grassmann distance metrics is not straightforward because each metric $d_{\mathrm{Gr}}$ has its own intrinsic scale and functional form. 
The standard conformal radius $R = \widehat{q}_{1-\alpha}$ (or the adaptive radius $R^* = \widehat{q}_{1-\alpha}^{\mathrm{n}} \widehat{\eta}(\mathbf{x}^*)$) is expressed in the native units of the chosen metric, meaning that a direct comparison conflates geometric size with arbitrary unit scaling.
To enable meaningful comparisons, standardized evaluation metrics are employed, discussed in the following.

To construct a universal geometric scale for target subspaces in $\mathrm{Gr}(p,n)$, we invert the native conformal radius of a prediction region back into a standardized Riemannian geodesic measure.
Because Grassmannian distance functions generally depend on the full vector of principal angles $(\theta_1,\dots,\theta_p)$ between the prediction and the true subspace, no universal scalar inversion exists for arbitrary dimensions. 
A comparable geometric scale can be obtained by assuming that all principal angles are equal, i.e., $\theta_1 = \dots = \theta_p = t$. 
Under this assumption, each native metric reduces to a function of $t$ and the native metric radius $R$ can be inverted into an \emph{equivalent geodesic radius} $R_{\mathrm{eq}} = \sqrt{p}\,t$.
Setting the native distance formulas equal to $R$, an explicit map is obtained for each Grassmannian distance metric:
\begin{subequations}
\label{eq:isotropic-inversions-general}
\begin{align}
\text{Geodesic:} \quad &R_{\mathrm{eq}} = R, \\
\text{Projection:} \quad &R_{\mathrm{eq}} = \sqrt{p} \arcsin\left(\frac{R}{\sqrt{p}}\right), \\
\text{Binet-Cauchy:} \quad &R_{\mathrm{eq}} = \sqrt{p} \arccos\left((1 - R^2)^{\frac{1}{2p}}\right), \\
\text{Procrustes:} \quad &R_{\mathrm{eq}} = \sqrt{p} \arccos\left(1 - \frac{R^2}{2p}\right).
\end{align}
\end{subequations}

The equivalent radius associates the native metric radius with the geodesic radius obtained under an equal-principal-angle reference configuration, $\theta_1 = \cdots = \theta_p = t.$ This defines an idealized geodesic ball
\begin{equation}
\mathcal{B}_{\mathrm{geo}}
\left(
\widehat{\mathcal{Y}},
R_{\mathrm{eq}}
\right),
\qquad
R_{\mathrm{eq}} = \sqrt{p}\,t.
\end{equation}
We then use the relative volume of this geodesic ball as a unit-free proxy for prediction-region size:
\begin{equation}
\label{eq:volume-fraction}
f_{\mathrm{vol}} = \frac{\operatorname{Vol}\left( \mathcal{B}_{\mathrm{geo}}
\left( \widehat{\mathcal{Y}}, R_{\mathrm{eq}} \right) \right)}{\operatorname{Vol}\left( \mathrm{Gr}(p,n) \right)}.
\end{equation}
For $p>1$, $f_{\mathrm{vol}}$ should be interpreted as a standardized geodesic-volume approximation rather than the exact volume of the original metric ball, since the equal-principal-angle construction provides only a scalar reference configuration. 
In this case, $f_{\mathrm{vol}}$ is computed numerically via the Weyl integration formula \cite{chikuse2003statistics}.
For $p=1$, there is a single principal angle and the formula is exact.

\subsection{Benchmark 1: Unit sphere}
\label{sec:sphere}
The first benchmark is adapted from the work of Shahbazi and Baheri \cite{shahbazi2026geometry}, where data are generated on the unit sphere $\mathbb{S}^2 = \left\{\mathbf{y} \in \mathbb{R}^3 : \lVert\mathbf{y}\rVert_2 = 1 \right\}$. 
The unit sphere coincides with the oriented Grassmann manifold $\mathrm{Gr}^+(1,3)$, where two antipodal points on the sphere are distinct. 
In contrast, on the standard (unoriented) Grassmann manifold $\mathrm{Gr}(1,3)$  those antipodal points are identified as the same subspace, i.e., they correspond to the same point on the manifold. 

\begin{figure}[t!]
\centering
\includegraphics[width=1.0\textwidth]{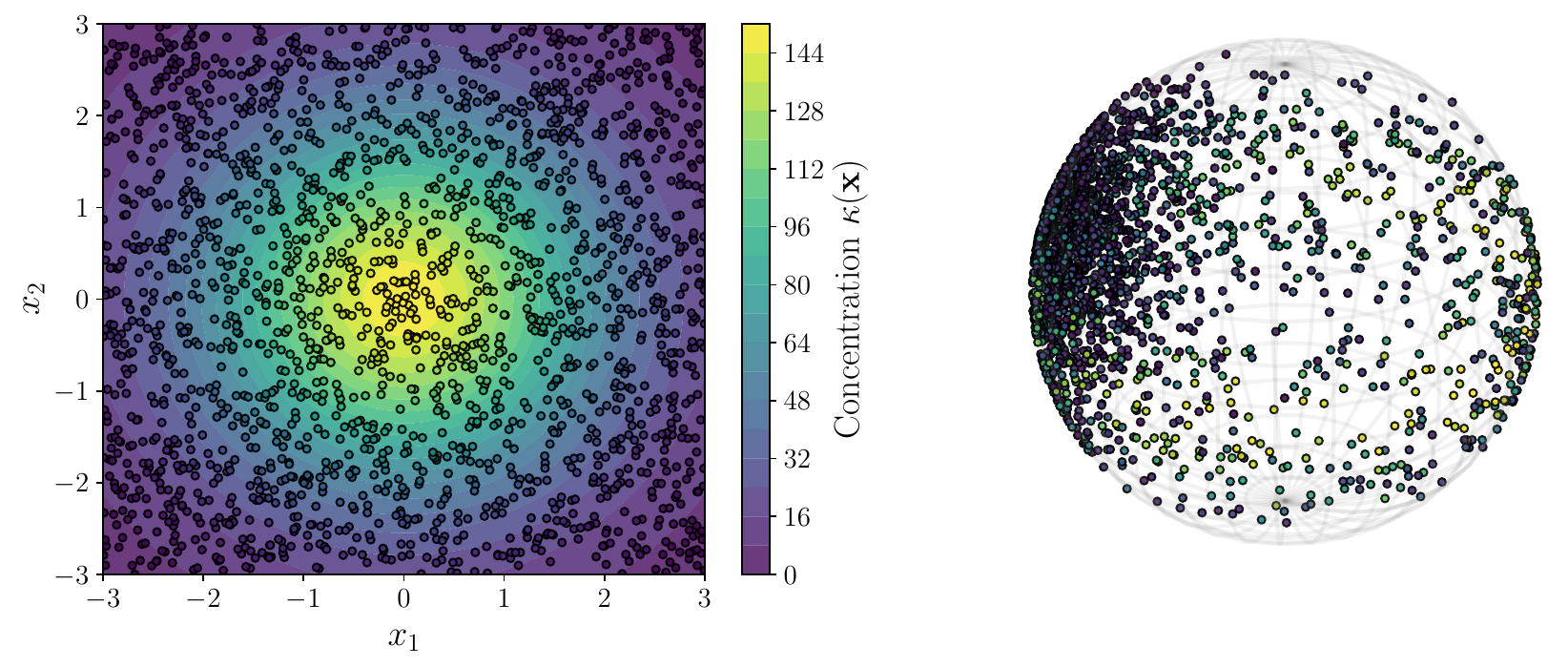}
\caption{Data generation for the unit-sphere benchmark. \textbf{Left:} Sampling of the input domain. \textbf{Right:} Responses on the unit sphere.}
\label{fig:sphere-data-generation}
\end{figure}

We sample $Q = 1200$ points with inputs $\mathbf{x} \in \left[-3,3\right]^2$ and responses $\mathbf{y} \sim \mathrm{vMF} \left(m(\mathbf{x}), \kappa(\mathbf{x})\right)$, where $\mathrm{vMF} \left(m(\mathbf{x}), \kappa(\mathbf{x})\right)$ is a von Mises-Fisher distribution with mean function
\begin{equation}
\label{eq:S2-mean-function}
m(\mathbf{x}) = \left(\sin{\left(\vartheta(\mathbf{x})\right)} \cos{\left(\phi(\mathbf{x})\right)}, \sin{\left(\vartheta(\mathbf{x})\right)} \sin{\left(\phi(\mathbf{x})\right)}, \cos{\left(\vartheta(\mathbf{x})\right)}\right),
\end{equation}
with $\vartheta\left(\mathbf{x}\right) = \frac{\pi}{2} + 0.3 \tanh{\left(x_1\right)}$ and $\phi(\mathbf{x}) = \pi \tanh{\left(x_2\right)}$, and concentration function 
\begin{equation}
\label{eq:S2-concentration-function}
\kappa(\mathbf{x}) = 3 + 147 \exp{\left( - \frac{\| \mathbf{x}\|^2}{4}\right)}.
\end{equation}
As also illustrated in Figure~\ref{fig:sphere-data-generation}, the responses $\mathbf{y} \in \mathbb{S}^2$ exhibit strong heteroscedasticity across the input domain. 

In addition to the four Grassmannian distance metrics defined in \eqref{eq:grassmann-distances}, we additionally use the native $\mathbb{S}^2$ geodesic distance, defined as
\begin{equation}
\label{eq:S2-geodesic-distance}
d_{\mathbb{S}^2}\left(\mathbf{y}_1, \mathbf{y}_2\right) = \arccos\left(\langle \mathbf{y}_1, \mathbf{y}_2 \rangle\right).
\end{equation}
This distance metric measures the length of the shortest great-circle arc connecting the two points and respects orientation, i.e., $d_{\mathbb{S}^2}\left(\mathbf{y}_1, \mathbf{y}_2\right) \neq d_{\mathbb{S}^2}\left(\mathbf{y}_1, -\mathbf{y}_2\right)$ in general.
In contrast, any of the Grassmannian distance metrics satisfies $d_{\mathrm{Gr}}\left(\mathbf{y}_1, \mathbf{y}_2\right) = d_{\mathrm{Gr}}\left(\mathbf{y}_1, -\mathbf{y}_2\right)$.
To ensure a unified, directly comparable, and unit-free comparison of prediction region sizes, we compute the volume fraction $f_{\mathrm{vol}}$ by representing each prediction region by a spherical cap of equivalent geodesic radius $R_{\mathrm{eq}}$ (defined in Section~\ref{sec:cross-metric-standardization}) and normalizing its area by the total sphere area.
The volume fraction~\eqref{eq:volume-fraction} then simplifies to
\begin{equation}
\label{eq:volume-fraction-sphere}
f_{\mathrm{vol}} = \frac{1 - \cos R_{\mathrm{eq}}}{2}.
\end{equation}
Note that, in this particular case, $f_{\mathrm{vol}}$ corresponds to a (normalized) surface area.

As an additional baseline metric, we use the naive $L_\infty$ distance on the spherical coordinate chart, which produces a \gls{cp} region that forms an axis-aligned coordinate rectangle instead of a spherical cap. 
The area of that \gls{cp} region on $\mathbb{S}^2$ is calculated analytically as $4R\sin(R)\sin(\vartheta_0)$, where $R$ is the conformal radius (half-width of the rectangle) and $\vartheta_0 = \arccos(\widehat{y}_3)$ is the polar angle of the predicted center $\widehat{\mathbf{y}}$. 
This area is normalized by the total surface area of the sphere to obtain the corresponding comparable volume fraction.

\begin{table}[b!]
\caption{Performance of standard and adaptive CP on the unit sphere, using different distance metrics (naive $L_\infty$, $\mathbb{S}^2$ geodesic, Grassmannian). Results for $\alpha=0.1$ over 500 seeds.}
\label{tab:sphere}
\centering
\begin{tabular}{lccccc}
\toprule
Distance metric & Avg. Cov. ($\uparrow$) & Worst Cov. ($\uparrow$) & Bin Cov. Std. ($\downarrow$) & Equiv. Radius ($\downarrow$) & Vol. Frac. ($\downarrow$) \\
\midrule
$L_{\infty}$ (standard) & $0.902 \pm 0.022$ & $0.7390$ & $0.0867$ & N/A & $0.07222$ \\
$L_{\infty}$ (adaptive) & $0.901 \pm 0.022$ & $0.8142$ & $0.0527$ & N/A & $0.06920$ \\
\midrule
$\mathbb{S}^2$ geodesic (standard) & $0.901 \pm 0.022$ & $0.7251$ & $0.0924$ & $0.5257$ & $0.06774$ \\
$\mathbb{S}^2$ geodesic (adaptive) & $0.900 \pm 0.022$ & $0.8096$ & $0.0544$ & $0.4997$ & $0.06609$ \\
\midrule 
Gr. geodesic (standard) & $0.901 \pm 0.022$ & $0.7251$ & $0.0924$ & $0.5257$ & $0.06774$ \\
Gr. geodesic (adaptive) & $0.900 \pm 0.022$ & $0.8096$ & $0.0544$ & $0.4997$ & $0.06609$ \\
\midrule
Projection (standard) & $0.901 \pm 0.022$ & $0.7251$ & $0.0924$ & $0.5257$ & $0.06774$ \\
Projection (adaptive) & $0.901 \pm 0.022$ & $0.8139$  & $0.0523$ & $0.5049$ & $0.06838$ \\
\midrule
Procrustes (standard) & $0.901 \pm 0.022$ & $0.7251$ & $0.0924$ & $0.5257$ & $0.06774$ \\
Procrustes (adaptive) & $0.900 \pm 0.022$ & $0.8102$ & $0.0540$ & $0.5007$ & $0.06650$ \\
\midrule
Binet-Cauchy (standard) & $0.901 \pm 0.022$ & $0.7251$ & $0.0924$ & $0.5257$ & $0.06774$ \\
Binet-Cauchy (adaptive) & $0.901 \pm 0.022$ & $0.8139$ & $0.0523$ & $0.5049$ & $0.06838$ \\
\bottomrule
\end{tabular}
\end{table}

Table~\ref{tab:sphere} presents numerical results over $500$ seeds for $\alpha=0.1$. 
The initial dataset ($Q = 1200$ data points) is partitioned to $M=480$ training data points, $J=360$ calibration data points, and $L=360$ test data points.
A $k$-nearest neighbors regression algorithm \cite{song2017efficient} with $k=20$ is used for both the base model $\widehat{\mu}$ and the auxiliary model $\widehat{\gamma}$ that is used within the difficulty estimator $\widehat{\eta}$.  
To estimate the coverage standard deviation over equal-frequency difficulty bins with different prediction difficulty, we use the true difficulty $\eta(\mathbf{x}) = 1 / \kappa(\mathbf{x})$ and $N_\mathrm{bins}=6$.
We report the empirical marginal coverage (including its standard deviation across trials), the worst-bin coverage, the standard deviation of coverage across bins, the equivalent radius, and the volume fraction.

Certain points concerning the results presented in Table~\ref{tab:sphere} require clarification:
\begin{itemize}
\item The results of standard (non-adaptive) \gls{cp} are identical across all Grassmannian distance metrics. 
On $\mathbb{S}^2$ ($p=1$), there is a single principal angle $\theta \in \left[0, \pi/2\right]$ between the predicted and true response. 
Additionally, every Grassmannian distance is a strictly monotone function of $\theta$.
Therefore, the sorting order of the corresponding non-conformity scores does not change, irrespective of the metric used.
That is, if point $\mathbf{y}_A$ is geometrically farther from its prediction compared to point $\mathbf{y}_B$, then $s_A > s_B$ regardless of the Grassmann distance metric.
Consequently, the exact same data point is selected as $\widehat{q}_{1-\alpha}$, which in turn translates to the same equivalent radius, volume fraction, and empirical coverage results.
\item The results of the $\mathbb{S}^2$ and the Grassmannian geodesic metrics are identical. 
These two distances differ only in the absolute value used in the Grassmannian geodesic metric. 
Therefore, their evaluations differ only if $\langle \mathbf{y}, \widehat{\mathbf{y}} \rangle < 0$, i.e., when the true and predicted responses lie on opposite hemispheres. 
Due to the accuracy of the base regression model $\widehat{\mu}$, this never occurs.
\item The results corresponding to the projection and Binet-Cauchy metrics are identical, as both distances reduce to the same expression for $p=1$.
\item For the projection and Binet-Cauchy metrics, the volume fraction slightly increases when adaptive \gls{cp} is employed, despite the decrease in the equivalent radius. 
This paradox is a consequence of Jensen's inequality acting on a highly skewed radius distribution. 
In adaptive \gls{cp}, the equivalent radius $R_{\mathrm{eq}} = \arcsin\left(\widehat{q}_{1-\alpha}^{\mathrm{n}} \widehat{\eta}(\mathbf{x})\right)$ is reduced for the majority of points in low-difficulty regions, resulting in an overall reduction (on average). 
However, because the $\arcsin$ function is nonlinear and convex on the relevant radius range, linearly scaling the score in high-difficulty regions results in a nonlinear inflation of $R_{\mathrm{eq}}$ for a minority of hard-to-predict test points. 
The map from equivalent radius to volume fraction, $f_{\mathrm{vol}} = (1 - \cos R_{\mathrm{eq}})/2$, is convex over the range of radii encountered in this experiment. 
Consequently, the volume penalty incurred by the outliers outweighs the volume savings from the majority of easy-to-predict points. 
This results in a slightly increased volume fraction, even though the equivalent radius is reduced.
\end{itemize}

\begin{figure}[t!]
\centering
\includegraphics[width=1.0\textwidth]{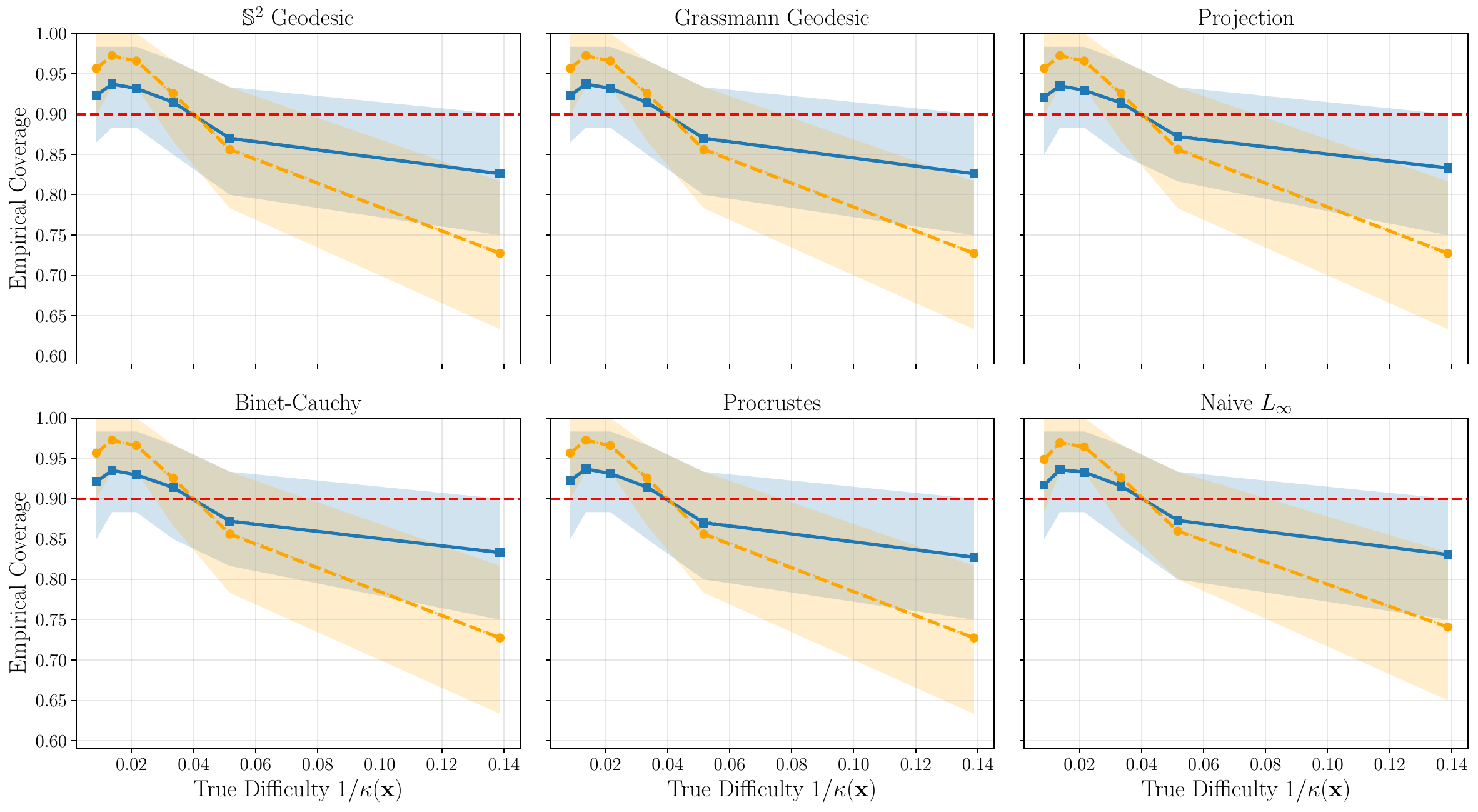}%
\vspace{0em}
\includegraphics[width=1\textwidth]{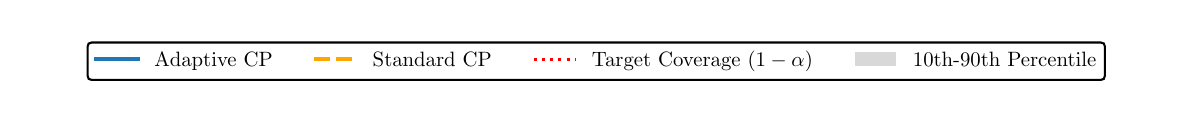}
\caption{Empirical coverage provided by standard and adaptive \gls{cp} over the six difficulty bins, for all considered distance metrics. Average results and percentiles over 500 seeds are shown.}
\label{fig:standard-vs-adaptive-CP-sphere}
\end{figure}

All \gls{cp} variants achieve the target (marginal) coverage regardless of distance metric or adaptivity, with a relatively small standard deviation over trials.
Locally adaptive \gls{cp} consistently improves worst-bin coverage, standard deviation over difficulty bins, equivalent radius, and volume fraction. 
The only exception is the volume fraction for the projection and Binet-Cauchy metrics, which slightly increases due to the reasons explained above.
Figure~\ref{fig:standard-vs-adaptive-CP-sphere} illustrates the underlying reason for these improvements.
While the empirical coverage of standard \gls{cp} drops sharply as prediction difficulty increases, adaptive \gls{cp} effectively mitigates this decline, yielding much more uniform coverage across difficulty bins.

\begin{figure}[t!]
\centering
\includegraphics[width=0.8\textwidth]{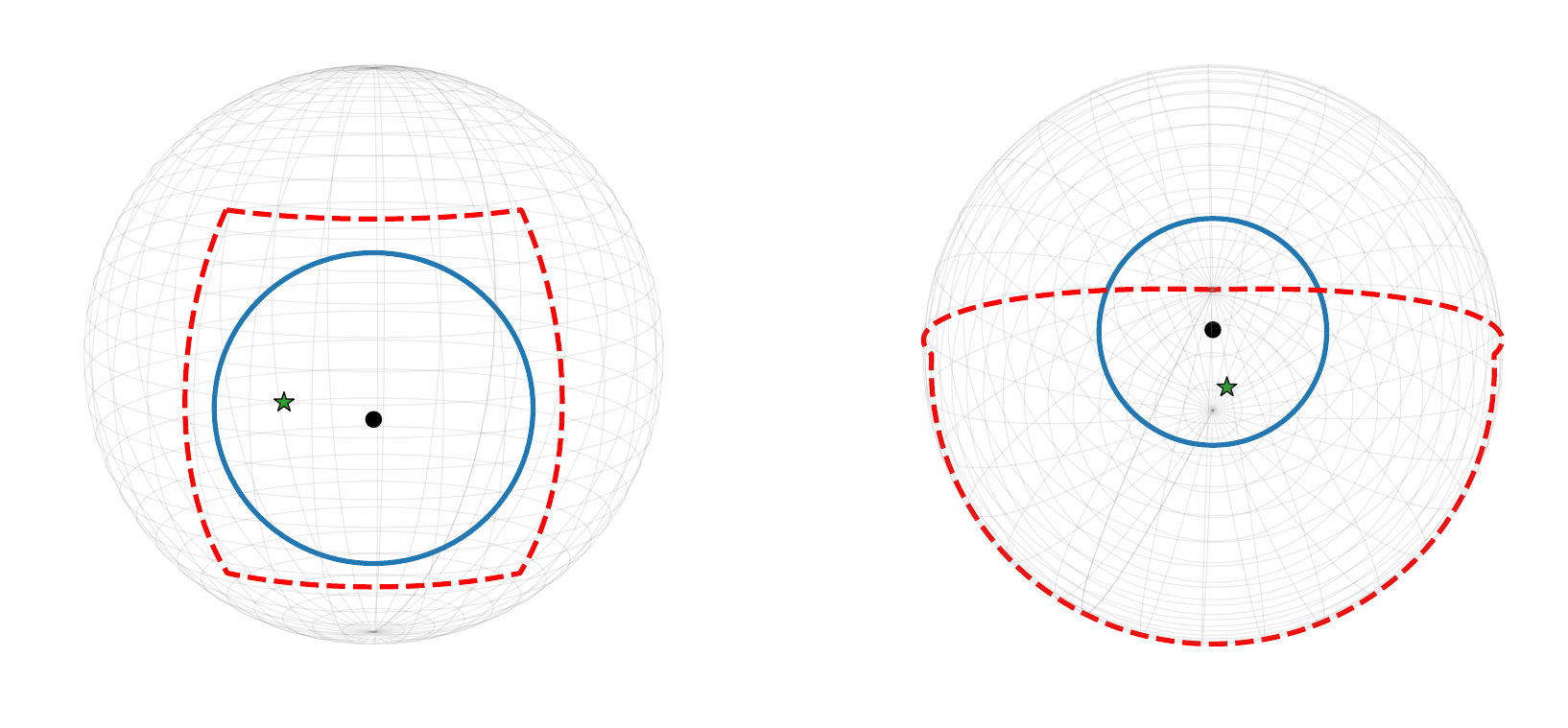}%
\vspace{-1.5em}
\includegraphics[width=0.8\textwidth]{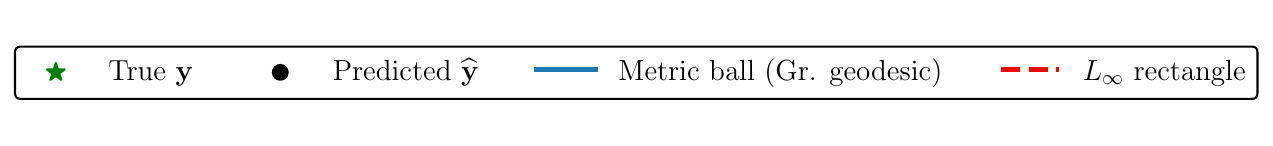}
\caption{\textbf{Left:} $L_\infty$ rectangle versus metric ball (here, spherical cap) based on the Grassmannian geodesic distance for a point near the equator. The rectangle covers a significantly larger area for the same empirical coverage. \textbf{Right:} Distortion of the $L_\infty$ rectangle near the poles. }
\label{fig:ball-vs-rectangle}
\end{figure}

The results also highlight the necessity of employing basis-invariant, geometry-aware Grassmannian metrics. 
Compared to the Grassmannian metrics, the naive $L_\infty$ metric yields a notably higher volume fraction in both standard and adaptive \gls{cp} variants, thus resulting in a loss of statistical efficiency. 
This is visually confirmed in Figure~\ref{fig:ball-vs-rectangle}, where the prediction regions (spherical caps) generated by the Grassmannian metrics cover a much smaller area than the prediction regions (rectangles) produced by the $L_\infty$ metric. 
The latter also suffer from significant geometric distortion for predictions near the poles, as shown in the same figure.

\subsection{Benchmark 2: Rotating plane}
\label{sec:rot-plane}
As a second benchmark, we consider the Grassmannian $\mathrm{Gr}\left(2,4\right)$, each point of which is a plane through the origin of $\mathbb{R}^4$.
A subspace $\mathcal{Y} \in \mathrm{Gr}\left(2,4\right)$ is represented by an orthonormal matrix $\mathbf{Y} \in \mathbb{R}^{4 \times 2}$ with $\mathbf{Y}^\top \mathbf{Y} = \mathbf{I}_2$. 
For a scalar input $x \in \left[0, 2\pi\right]$, the orthonormal matrix that spans the corresponding subspace $\mathcal{Y}(x)$ is given by
\begin{equation}
\mathbf{Y}(x) = 
\begin{bmatrix}
\cos{(x)} & 0 \\
\sin{(x)} & 0 \\
0 & \cos{(x)} \\ 
0 & \sin{(x)}
\end{bmatrix}.
\end{equation} 
This subspace rotates at constant speed as $x$ increases. 
When $x$ increases by $\pi$, both spanning vectors flip sign, meaning that the (unoriented) subspace completes a full revolution and maps back onto itself.
After $2\pi$, the basis matrix $\mathbf{Y}(x)$ also returns to its original orientation.
 
Dropping the dependence on $x$ to ease the notation, a perturbed basis $\widetilde{\mathbf{Y}}$ is obtained as follows:
\begin{enumerate}
\item A random matrix $\mathbf{Z} \in \mathbb{R}^{4 \times 2}$ with \gls{iid} standard normal entries is drawn and projected onto the tangent space $\mathcal{T}_{\mathbf{Y}}$. The projection matrix is given as
\begin{equation}
\mathbf{T} = \left( \mathbf{I} - \mathbf{Y} \mathbf{Y}^\top \right) \mathbf{Z}.
\end{equation} 

\item An orthonormal basis $\mathbf{Y}_{\perp} \in \mathbb{R}^{4 \times 2}$ of the orthogonal complement is computed via QR decomposition. 
We then form $\mathbf{B} = \left(\mathbf{Y}_{\perp}\right)^\top \mathbf{T} \in \mathbb{R}^{2 \times 2}$ and its SVD $\mathbf{B} = \mathbf{V} \boldsymbol{\Sigma} \mathbf{W}^\top$. 

\item The scaled singular value matrix $\widetilde{\boldsymbol{\Sigma}} = \frac{\eta(x)}{\|\boldsymbol{\Sigma}\|_{\mathrm{F}}} \boldsymbol{\Sigma}$ is computed, where
\begin{equation}
\eta(x) = 0.2 + 0.8 \sin^2{(x)}
\end{equation} 
is a noise function that peaks at $x = \pi/2$ and $x = 3\pi/2$ and is minimal at $x=0$, $x=\pi$, and $x=2\pi$.

\item The perturbed orthonormal matrix is obtained as $\widetilde{\mathbf{Y}} = \mathbf{Y} \mathbf{W} \cos{\left( \widetilde{\boldsymbol{\Sigma}} \right)} + \mathbf{Y}_{\perp} \mathbf{V} \sin{\left( \widetilde{\boldsymbol{\Sigma}} \right)}$. 

\end{enumerate} 

Table~\ref{tab:rot-plane} presents the numerical results for $\alpha=0.1$ over 100 seeds, where the full dataset $\left(x^{(q)}, \widetilde{\mathbf{Y}}^{(q)}\right)_{q=1}^Q$ consists of $Q=1200$ data points, partitioned to $M=480$ training data, $J=360$ calibration data, and $L=360$ test data.
Similar to Section~\ref{sec:sphere}, the base model $\widehat{\mu}$ is a $k$-nearest neighbors regressor with $k=20$.  
The auxiliary regressor $\widehat{\gamma}$ used within the difficulty estimator $\widehat{\eta}$ is a Gaussian process model \cite{rasmussen2006gaussian} with a composite kernel consisting of a constant kernel, a periodic kernel, and a white noise kernel. 
The composite kernel is given as
\begin{equation}
k\left(x, x'\right) = \kappa_{\mathrm{scale}} \exp\left( -\frac{2 \sin^2\left(\frac{1}{2}\lvert x - x'\rvert\right)}{\ell^2} \right) + \kappa_{\mathrm{noise}} \delta\left(x, x'\right),
\end{equation}
where $\kappa_{\mathrm{scale}}$ and $\kappa_{\mathrm{noise}}$ are the variances of the smooth signal and white noise components, respectively, $\ell$ is the characteristic length-scale controlling the smoothness of the estimated noise envelope, and $\delta$ is the Kronecker delta function.
The hyper-parameters $\kappa_{\mathrm{scale}}$,  $\kappa_{\mathrm{noise}}$, and $\ell$ are optimized via the L-BFGS-B algorithm \cite{zhu1997algorithm} by maximizing the marginal log-likelihood of the log-transformed \gls{cv}-residuals.
To estimate the standard deviation of coverage across equal-frequency difficulty bins, we use the true difficulty $\eta(x) = 0.2 + 0.8 \sin^2(x)$ and $N_{\mathrm{bins}}=12$.

\begin{table}[b!]
\caption{Performance of standard and adaptive CP on the rotating plane, using different Grassmannian distance metrics. Results for $\alpha=0.1$ over 100 seeds.}
\label{tab:rot-plane}
\centering
\begin{tabular}{lccccc}
\toprule
Distance metric & Avg. Cov. ($\uparrow$) & Worst Cov. ($\uparrow$) & Bin Cov. Std. ($\downarrow$)  & Equiv. Radius ($\downarrow$) & Vol. Frac. ($\downarrow$) \\
\midrule
Gr. geodesic (standard) & $0.905 \pm 0.021$ & $0.5908$ & $0.1538$ & $1.0783$ & $0.25998$ \\
Gr. geodesic (adaptive) & $0.905 \pm 0.023$ & $0.7958$ & $0.0644$ & $0.7436$ & $0.14065$ \\
\midrule
Projection (standard) & $0.906 \pm 0.023$ & $0.5980$ & $0.1511$ & $0.9888$ & $0.19178$ \\
Projection (adaptive) & $0.905 \pm 0.022$ & $0.8200$ & $0.0521$ & $0.7025$ & $0.11168$ \\
\midrule
Procrustes (standard) & $0.905 \pm 0.022$ & $0.5932$ & $0.1527$ & $1.0601$ & $0.24509$ \\
Procrustes (adaptive) & $0.906 \pm 0.024$ & $0.8028$ & $0.0616$ & $0.7351$ & $0.13448$ \\
\midrule
Binet-Cauchy (standard) & $0.904 \pm 0.022$ & $0.5893$ & $0.1545$ & $1.1241$ & $0.30011$ \\
Binet-Cauchy (adaptive) & $0.903 \pm 0.023$ & $0.8117$ & $0.0568$ & $0.8092$ & $0.20853$ \\
\bottomrule
\end{tabular}
\end{table}

\begin{figure}[t!]
\centering
\includegraphics[width=1\textwidth]{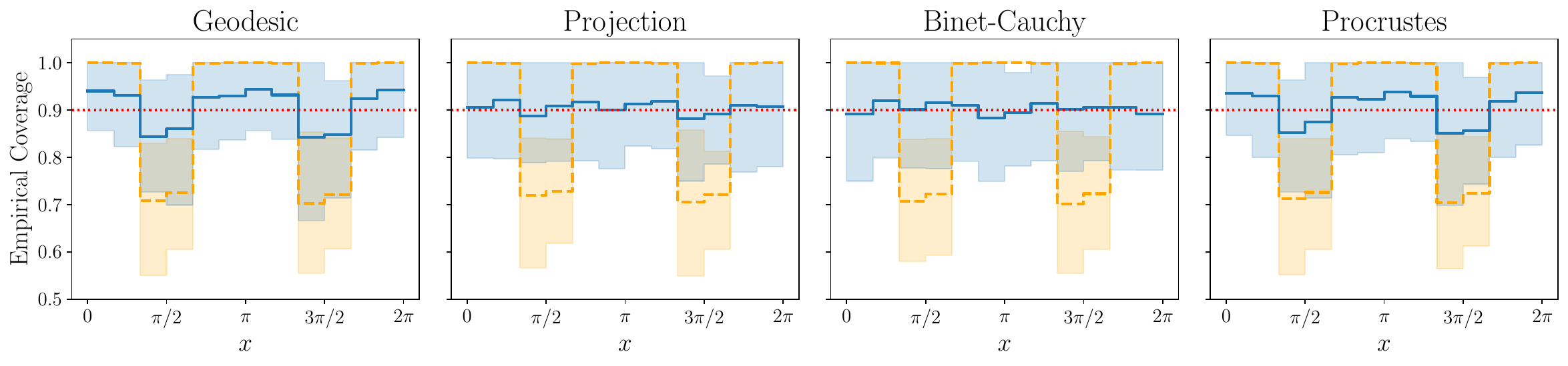}%
\vspace{-0.5em}
\includegraphics[width=1\textwidth]{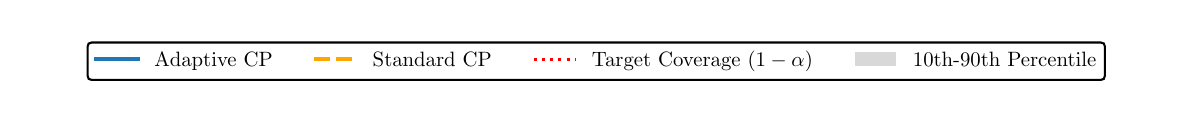}
\caption{Empirical coverage across the input domain (partitioned into $N_\mathrm{bins}=12$ difficulty bins) for the standard and adaptive \gls{cp} and for all Grassmannian distance metrics. Average results and percentiles over 100 seeds are shown.}
\label{fig:plane-coverage}
\end{figure}

Once again, regardless of distance metric or adaptivity, all \gls{cp} variants achieve the desired marginal coverage, with a small standard deviation over the 100 trials.
However, standard \gls{cp} exhibits severe empirical miscoverage across difficulty strata in this highly heteroscedastic setting. 
For all distance metrics, standard \gls{cp} exhibits a severe drop in the worst-bin coverage, falling to less than $60\%$.
Additionally, the high standard deviation of coverage across bins indicates a severe imbalance, meaning that low-noise regions are heavily over-covered, while high-noise regions are drastically under-covered.
The adaptive \gls{cp} approach mitigates this imbalance.
The worst-bin coverage improves substantially across all metrics to approximately $80\%$ or above, while the standard deviation across bins drops by a factor of nearly three.
Additionally, statistical efficiency is improved, as the average equivalent radius and volume fraction of the prediction regions are significantly reduced. 
This systematic failure of standard \gls{cp} and the mitigating effect of the adaptive approach are visually confirmed in Figure~\ref{fig:plane-coverage}.
Standard \gls{cp} shows a distinct drop in coverage specifically within the peak-noise regions ($x=\pi/2$ and $x=3\pi/2$), whereas adaptive \gls{cp} maintains a much more uniform coverage across the input domain, particularly when using the projection and Binet-Cauchy metrics.

\begin{figure}[t!]
\centering
\includegraphics[width=1\textwidth]{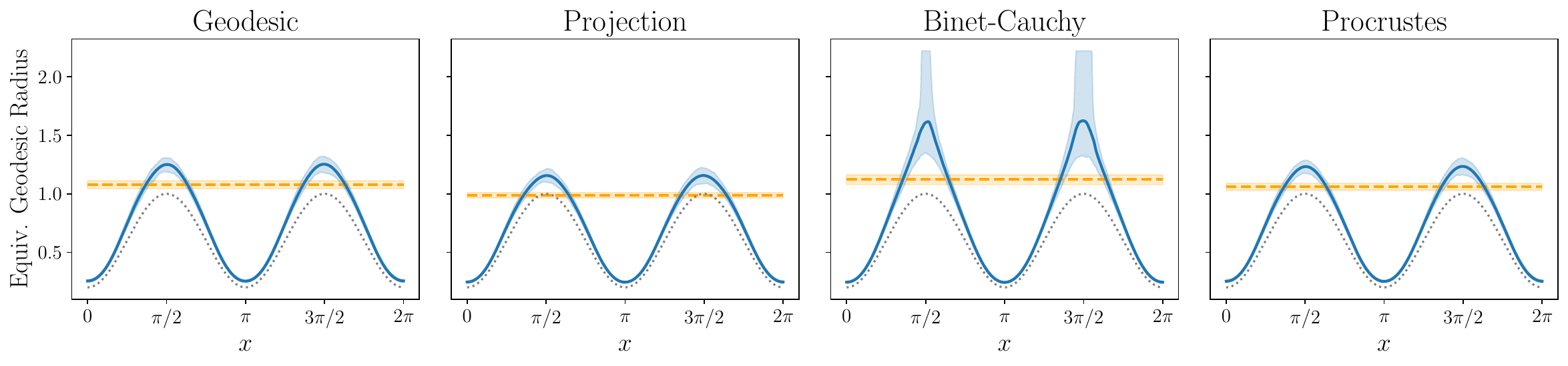}%
\vspace{-0.5em}
\includegraphics[width=1\textwidth]{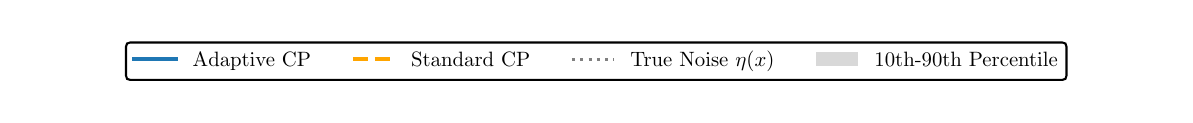}
\caption{Equivalent geodesic radius over the input domain for the standard and adaptive \gls{cp} and for all Grassmannian distance metrics. Average results and percentiles over the 100 seeds are shown. The true noise function over the input domain is shown as well.}
\label{fig:plane-equiv-radius}
\end{figure}

Comparing the different distance metrics on the Grassmannian, the projection distance provides the most favorable performance for this benchmark. 
It yields the highest worst-bin coverage and the lowest standard deviation over bins, while also producing the smallest \gls{cp} regions. 
In contrast, the Binet-Cauchy distance proves to be the most conservative metric, resulting in significantly larger equivalent radii and volumes.
This behavior is visually confirmed in Figure~\ref{fig:plane-equiv-radius}, which illustrates the equivalent geodesic radius across the input domain. 
The figure also demonstrates that the projection metric produces locally adaptive radii that track the true noise function more accurately than the other metrics, which explains its statistical efficiency advantage. 

\section{Application to airfoil inverse shape design}
\label{sec:airfoil}

In airfoil inverse design a prescribed aerodynamic response must be mapped back to a compatible geometry, such that the inferred airfoil shape produces the desired response. 
This inverse map can be non-unique and locally ill-conditioned, since different geometries may exhibit similar aerodynamic characteristics. 
Quantifying uncertainty in the predicted geometry is therefore essential for assessing design reliability and robustness. 
Since the airfoil geometry admits a representation on the Grassmann manifold~\cite{grey2023separable, doronina2025}, the Grassmannian \gls{cp} framework introduced in Section~\ref{sec:acp-grassmann} is a natural option for quantifying this uncertainty directly on the manifold. 
The resulting prediction regions describe sets of plausible airfoil shapes associated with a prescribed aerodynamic response. 
We hypothesize that the size of prediction regions carries an engineering interpretation: small prediction regions correspond to aerodynamic conditions that constrain the geometry relatively well, whereas larger regions indicate greater ambiguity or sensitivity in the inverse-design map.

For each airfoil, the input covariate vector is the aerodynamic triad $(C_L, C_D, C_M)$, where $C_L$, $C_D$, and $C_M$ denote the lift, drag, and pitching-moment coefficients, respectively. 
These are evaluated at an angle of attack $\mathrm{AoA}=4^\circ$, which is the angle between the airfoil chord line and the incoming flow.
Each airfoil is represented by an orthonormal matrix
$\mathbf{Y}\in\mathbb{R}^{1001\times 2}$. 
The response therefore lies on $\mathrm{Gr}(2,1001)$ ($n=1001, p=2$), whose intrinsic dimension is
$2\times(1001-2)=1998$. 
However, it has been previously shown that only four dimensions are required to cover a reasonable engineering design space by using a four-dimensional section of the Grassmannian~\cite{doronina2025}. 
Thus, the approximations employed here are significantly more tractable by virtue of this dimension reduction.
Specifically, tangent principal components analysis~\cite{grey2023separable, doronina2025, fletcher2004principal} is used to learn the four dimensional domain from a curated database of airfoils. 
The same dataset is used in this study, which contains $Q = 8997$ airfoil samples in total.  
Using a random partition, we allocate $M = 4498$ samples to training, $J = 2249$ samples to calibration, and $L = 2250$ samples to testing.   
The base regressor is an inverse ridge model defined in Grassmann tangent coordinates~\cite{doronina2025}. 
Specifically, the Grassmann-valued training responses are mapped to the tangent space at a reference subspace, and ridge regression, akin to the structure revealed in ~\cite{doronina2025}, is used to map the complete aerodynamic covariate vector to the corresponding tangent coordinates. 
The predicted four-dimensional tangent vector is subsequently lifted to $\mathrm{Gr}(2,1001)$ using the Grassmann exponential map (see Section~\ref{sec:grassmann-regression}). 
The auxiliary model $\widehat{\gamma}$ used within the difficulty estimator $\widehat{\eta}$ is a $k$-nearest neighbors regressor with $k=20$.

We first evaluate the \gls{cp} regions directly on the Grassmann manifold. 
We set the significance level to $\alpha=0.1$, corresponding to a target coverage level of $1-\alpha=0.9$. 
To assess sensitivity to the random data partition, we repeat the experiment over 100 independent shuffles of the full dataset,  using the same data partition for the training, calibration, and test splits in each repetition.  
Table~\ref{tab:airfoil-grassmann-performance} reports predictive performance directly on $\mathrm{Gr}(2,1001)$ for standard and locally adaptive \gls{cp} using each of the four Grassmannian distance metrics. 
Note that we do not report a Grassmannian volume-fraction estimate, as the obtained numerical values are negligible due to the high intrinsic dimension of $\mathrm{Gr}(2,1001)$.
Both standard and adaptive \gls{cp} attain average marginal coverage similar to the nominal target of $0.9$.
In this case, performance is largely insensitive to the selected Grassmannian distance.
Adaptive \gls{cp} results in better uniformity across difficulty bins, however, this does not translate to tighter prediction regions. 
On the contrary, the equivalent radius slightly increases across all distance metrics.
Worst-bin coverage improves, but only slightly.
Note that $N_{\mathrm{bins}}=6$ equal-frequency difficulty bins are used, based on the estimated difficulty provided by $\widehat{\eta}(\mathbf{x})$.
Based on these results, in the following, we only use the adaptive \gls{cp} variant with the Grassmannian geodesic distance metric. 
The results obtained with the other basis-invariant distance metrics are similar.

\begin{table}[b!]
\centering
\caption{Performance of standard and adaptive \gls{cp} on the airfoil inverse shape-design problem.
Results for $\alpha=0.1$ and $100$ random splits of the original dataset.
For each split, the bin-coverage standard deviation is computed separately using equal-frequency partitions of $C_L$, $C_D$, and $C_M$ and then averaged
over the three coefficient orderings. 
Worst-bin coverage is the minimum coverage over all coefficient bins and orderings, averaged across splits.}
\label{tab:airfoil-grassmann-performance}
\begin{tabular}{lccccc}
\toprule
Distance metric
& Avg. Cov. ($\uparrow$)
& Worst Cov. ($\uparrow$)
& Bin Cov. Std. ($\downarrow$)
& Equiv. Radius ($\downarrow$)\\
\midrule
Gr. geodesic (standard)
& $0.901 \pm 0.008$ & $0.8414$ & $0.0400$ & $0.1678$ \\
Gr. geodesic (adaptive)
& $0.901 \pm 0.009$ & $0.8453$ & $0.0267$ & $0.1687$ \\
\midrule
Projection (standard)
& $0.901 \pm 0.008$ & $0.8414$ & $0.0400$ & $0.1674$ \\
Projection (adaptive)
& $0.901 \pm 0.008$ & $0.8454$ & $0.0269$ & $0.1683$ \\
\midrule
Procrustes (standard)
& $0.901 \pm 0.008$ & $0.8414$ & $0.0400$ & $0.1677$ \\
Procrustes (adaptive)
& $0.901 \pm 0.009$ & $0.8453$ & $0.0268$ & $0.1686$ \\
\midrule
Binet--Cauchy (standard)
& $0.901 \pm 0.008$ & $0.8414$ & $0.0400$ & $0.1680$ \\
Binet--Cauchy (adaptive)
& $0.901 \pm 0.009$ & $0.8454$ & $0.0269$ & $0.1690$ \\
\bottomrule
\end{tabular}
\end{table}

\begin{figure}[t!]
    \centering
    \begin{subfigure}{1\textwidth}
        \centering
        \includegraphics[width=\linewidth]{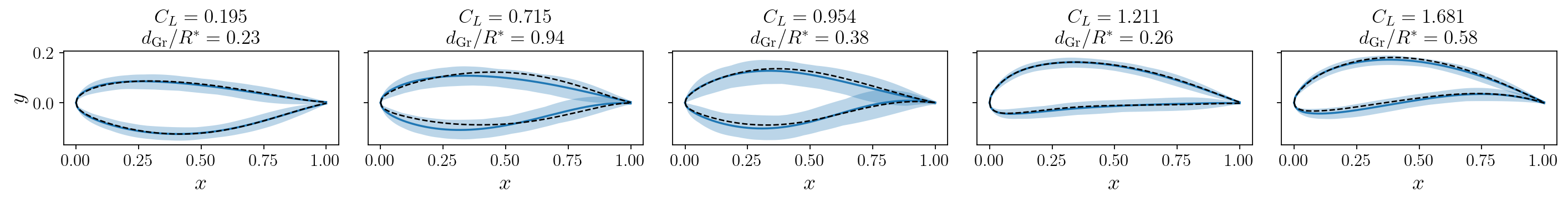}
        \caption{$C_L$-ordered validation examples.}
        \label{fig:validation_predictions_airfoil_cl}
    \end{subfigure}
    \begin{subfigure}{1\textwidth}
        \centering
        \includegraphics[width=\linewidth]{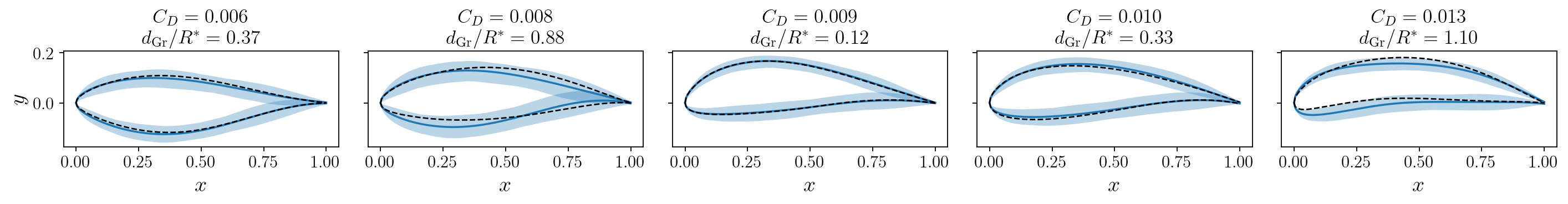}
        \caption{$C_D$-ordered validation examples.}
        \label{fig:validation_predictions_airfoil_cd}
    \end{subfigure}
    \begin{subfigure}{1\textwidth}
        \centering
        \includegraphics[width=\linewidth]{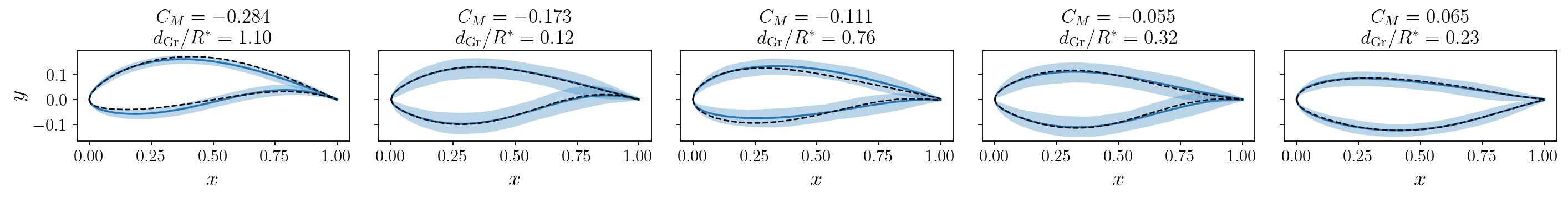}
        \caption{$C_M$-ordered validation examples.}
        \label{fig:validation_predictions_airfoil_cm}
    \end{subfigure}
    \begin{subfigure}{0.8\textwidth}
        \centering
        \includegraphics[width=\linewidth]{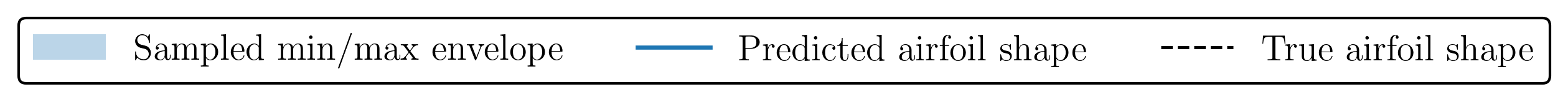}
        \label{fig:validation_predictions_airfoil_legend}
    \end{subfigure}
    \caption{Validation predictions of the conformalized Grassmannian regression model mapped to physical airfoil coordinates.
    Each panel shows the predicted airfoil shape, the pointwise min/max envelopes obtained from airfoil shapes sampled from the Grassmannian \gls{cp} region, and the true airfoil shape. 
    The quantity $d_{\mathrm{Gr}}/R^*$ is the ratio of the Grassmannian prediction error to the conformal radius, where values not exceeding one indicate coverage.  
    Every prediction uses the complete aerodynamic covariate vector $\mathbf{x}=(C_L,C_D,C_M)$.
    }
    \label{fig:validation_predictions_airfoil}
\end{figure}

Next, we visualize the prediction uncertainty in physical airfoil coordinates.
To that end, we sample subspace representations within the Grassmannian \gls{cp} region $\widehat{C}_{1-\alpha}(\mathbf{x}^*)$ using the base regression model, map the sampled subspaces back to airfoil shapes, and construct upper- and lower-surface bands around the predicted mean airfoil. 
These results are presented in Figure~\ref{fig:validation_predictions_airfoil}, where the displayed bands correspond to the pointwise minimum and maximum
of the physical-coordinate deviations obtained from $1000$ samples within the prediction region.
For each validation plot, five cases spanning the $C_L$, $C_D$, and $C_M$ ranges are selected from the held-out test set, where every prediction is conditioned on the complete covariate vector $\mathbf{x}=(C_L,C_D,C_M)$
Additionally, we use as diagnostic the ratio of the Grassmann prediction error to the \gls{cp} radius, $d_{\mathrm{Gr}}/R^*$. 
Ratios below one indicate coverage on the Grassmann manifold, whereas ratios larger than one indicate miscoverage.
It must be noted that the displayed upper- and lower-surface bands are only visual summaries induced by the Grassmannian \gls{cp} region and not \gls{cp} regions themselves. 
In particular, the reported min/max envelopes summarize the spread of the mapped samples at each airfoil coordinate. 
Conformal validity is associated only with membership of the true Grassmann-valued response in $\widehat{C}_{1-\alpha}(\mathbf{x}^*)$.
Therefore, there is no formal guarantee regarding the containment of the physical airfoil shape within the envelope for $d_{\mathrm{Gr}}/R^* \leq 1$, respectively lack of containment for $d_{\mathrm{Gr}}/R^* > 1$.
Nonetheless, we observe that there is close alignment in practice.

The predicted shapes generally follow the observed airfoil geometries closely, while the width of the physical-coordinate bands varies significantly across the coefficient ranges. 
This variation is attributed to the use of adaptive \gls{cp} on the Grassmann manifold, meaning that the locally adaptive prediction regions on the Grassmannian result in local envelope variations in physical coordinates. 
Figure~\ref{fig:validation_predictions_airfoil_cl} shows the validation examples ordered by $C_L$. 
All five representative cases are covered by their Grassmannian prediction regions, as indicated by the distance-to-radius ratios.
The true airfoil shapes are also contained within the envelopes, where the case $C_L=0.715$ has a ratio very close to one and the corresponding true airfoil almost exceeds the envelope. 
The $C_D$-ordered cases are shown in Figure~\ref{fig:validation_predictions_airfoil_cd}, where one instance is not covered ($C_D=0.013$). 
Figure~\ref{fig:validation_predictions_airfoil_cm} orders the cases by $C_M$, where again one representative case is not covered ($C_M=-0.284$). 
For both miscoverage cases, the corresponding envelopes do not fully contain the true airfoil shape.
Note that these misses are to be expected, as the selection of representative cases deliberately targets the extremes of the covariate range, where coverage is lowest.

\section{Conclusions}
\label{sec:conclusion}

This work introduced a locally adaptive conformal prediction framework for regression problems with responses residing on Grassmann manifolds. 
Rather than quantifying uncertainty in an ambient Euclidean representation, the proposed approach constructs prediction regions directly in the space of subspaces using basis-invariant Grassmannian distances. 
The resulting conformal prediction regions take the form of metric balls centered at the predicted subspace, with radii calibrated from normalized geometric non-conformity scores. 
A difficulty estimator learned from cross-validation log-residuals allows these regions to adapt to input-dependent variations in prediction difficulty while preserving the finite-sample marginal coverage guarantee of split conformal prediction under exchangeability. 

The numerical experiments demonstrate the importance of using geometry-aware non-conformity scores and local adaptation for regression on the Grassmann manifold. 
Across the synthetic benchmarks, both standard and adaptive conformal prediction attain the prescribed marginal coverage, irrespective of the employed distance metric. 
However, the non-adaptive regions can substantially under-cover difficult portions of the input space while over-covering easier regions. 
Scaling the Grassmannian residuals by a local difficulty estimate produces more uniform empirical coverage across difficulty levels and, in the strongly heteroscedastic rotating-plane problem (see Section~\ref{sec:rot-plane}), significantly reduces the average size of the prediction regions. 
Additionally, as shown in the unit sphere benchmark (see Section~\ref{sec:sphere}), the prediction regions resulting from the Grassmannian distance metrics are much more statistically efficient than Euclidean prediction regions.
The results further show that the choice of Grassmannian metric can also affect statistical efficiency, even if marginal validity is retained.  

The airfoil inverse-design problem further demonstrates the applicability of the framework to an engineering setting in which the predicted quantity is a geometric object and the inverse map can exhibit non-uniqueness and locally varying sensitivity. By representing airfoil shapes as Grassmann-valued responses, the method produces adaptive geometric uncertainty regions associated with prescribed aerodynamic coefficients. 
On the held-out test dataset, the resulting conformal predictor achieves empirical coverage close to the nominal target while providing input-dependent regions that can be mapped back to interpretable uncertainty bands in physical airfoil coordinates. 

More broadly, the results indicate that conformal prediction can be formulated directly in an intrinsic response space without sacrificing its distribution-free marginal validity. 
This separation between validity and geometry is important, as exchangeability determines the conformal coverage guarantee, whereas the choice of metric, base regressor, and difficulty estimator governs the shape, local adaptation, and statistical efficiency of the resulting prediction regions. 
The framework therefore provides a general approach for uncertainty quantification when the meaningful output is a subspace rather than an arbitrary Euclidean representation.

Several directions remain open. 
The equivalent-radius and volume measures were used in this work to provide a practical normalization for comparing different Grassmannian metrics, but their extension to exact anisotropic prediction-set volumes in higher-dimensional Grassmannians would enable more precise efficiency comparisons. 
Locally anisotropic conformal regions whose geometry adapts not only in radius but also along different tangent directions, may further improve efficiency when uncertainty is strongly directional. 
Other extensions include alternative estimators of local geometric difficulty, data-driven selection or combination of Grassmannian metrics, and conformal methods designed for weaker forms of exchangeability or distribution shift. 
Finally, extending the framework beyond Grassmann manifolds to broader classes of Riemannian and quotient manifolds would provide a general foundation for distribution-free uncertainty quantification of structured geometric outputs.

\paragraph{Funding statement:}
The work of D.G.~Giovanis was supported by the Defense Advanced Research Projects Agency (DARPA) under the COMPASS program (Agreement No. HR0011-25-3-0221).

\paragraph{Author contribution statement:}
All authors contributed to Conceptualization, Methodology, Formal Analysis, Writing -- original draft, and Writing -- review \& editing.
D.~Loukrezis and D.G.~Giovanis additionally contributed to Software, Investigation, Validation, and Visualization.

\paragraph{Data, software, and reproducibility statement:}
The software and data used to generate the numerical results reported in this work are available from the authors upon reasonable request.

\paragraph*{Conflict of interest statement:}
The authors have nothing to disclose.

\paragraph{Disclosure and declaration of AI use:}
During the preparation of this work, the authors used generative AI tools (various models of OpenAI and Anthropic) for software development, proofreading, and text-editing. 
All AI-generated outputs have been reviewed and edited by the authors, who take full responsibility for the content of this work.

\paragraph{Disclaimer:}
This publication was prepared as an account of a DARPA COMPASS collaboration. 
It is intended to capture external perspectives related to NIST technical work in applied and computational mathematics. 
External perspectives are included from academia, government, and other organizations. 
The opinions, recommendations, findings, and conclusions stated in this work do not necessarily reflect the views or policies of NIST or the United States Government.

\bibliographystyle{unsrt}
\bibliography{references}

\end{document}